\documentclass[10pt,twocolumn,letterpaper]{article}

\usepackage{cvpr}              

\usepackage{graphicx}
\usepackage{amsmath}
\usepackage{amssymb}
\usepackage{booktabs}
\usepackage{multirow}
\usepackage{tabularx}
\usepackage{appendix}

\usepackage[accsupp]{axessibility}  
\usepackage{algorithm}
\usepackage[noend]{algpseudocode}
\usepackage[pagebackref,breaklinks,colorlinks]{hyperref}

\usepackage[capitalize]{cleveref}
\crefname{section}{Sec.}{Secs.}
\Crefname{section}{Section}{Sections}
\Crefname{table}{Table}{Tables}
\crefname{table}{Tab.}{Tabs.}

\usepackage{cvpr}

\newcommand{\PAR}[1]{\vskip4pt \noindent{\bf #1~}}

\def\cvprPaperID{6567} 
\def\confName{CVPR}
\def\confYear{2023}

\begin{document}

\title{Privacy-Preserving Representations are not Enough: Recovering Scene Content from Camera Poses}

\author{Kunal Chelani$^{1}$
\quad
Torsten Sattler$^{2}$
\quad
Fredrik Kahl$^{1}$
\quad
Zuzana Kukelova$^{3}$\\
$^{1}$ Chalmers University of Technology \\
$^{2}$ Czech Institute of Informatics, Robotics and Cybernetics, Czech Technical University in Prague \\
$^{3}$ 
Visual Recognition Group,
Faculty of Electrical Engineering, Czech Technical University in Prague \\
{\tt\small \{chelani,fredrik.kahl\}@chalmers.se \, torsten.sattler@cvut.cz \, kukelzuz@fel.cvut.cz}
}


\maketitle

\begin{abstract}

Visual localization is the task of estimating the camera pose from which a given image was taken and is central to several 3D computer vision applications. With the rapid growth in the popularity of AR/VR/MR devices and cloud-based applications, privacy issues are becoming a very important aspect of the localization process. Existing work on privacy-preserving localization aims to defend against an attacker who has access to a cloud-based service. In this paper, we show that an attacker can learn about details of a scene without any access by simply querying a localization service. The attack is based on the observation that modern visual localization algorithms are robust to variations in appearance and geometry. While this is in general a desired property, it also leads to algorithms localizing objects that are similar enough to those present in a scene. An attacker can thus query a server with a large enough set of images of objects, \eg, obtained from the Internet, and some of them will be localized. The attacker can thus learn about object placements from the camera poses returned by the service (which is the minimal information returned by such a service). In this paper, we develop a proof-of-concept version of this attack and demonstrate its practical feasibility. The attack does not place any requirements on the localization algorithm used, and thus also applies to privacy-preserving representations. Current work on privacy-preserving representations alone is thus insufficient. 
\end{abstract}

\section{Introduction}
Visual localisation refers to the problem of estimating the camera pose of a given image in a known scene. 
It is a core problem in several 3D computer vision applications, including 
self-driving cars~\cite{Geppert2019ICRA,Heng2019ICRA} and other autonomous robots~\cite{Wendel11ICRA}, and Augmented Reality~\cite{Middelberg2014ECCV,Lynen2020IJRR,Castle08ISWC}.  

\begin{figure*}[t!]
    \centering
    \includegraphics[width=0.90\linewidth]{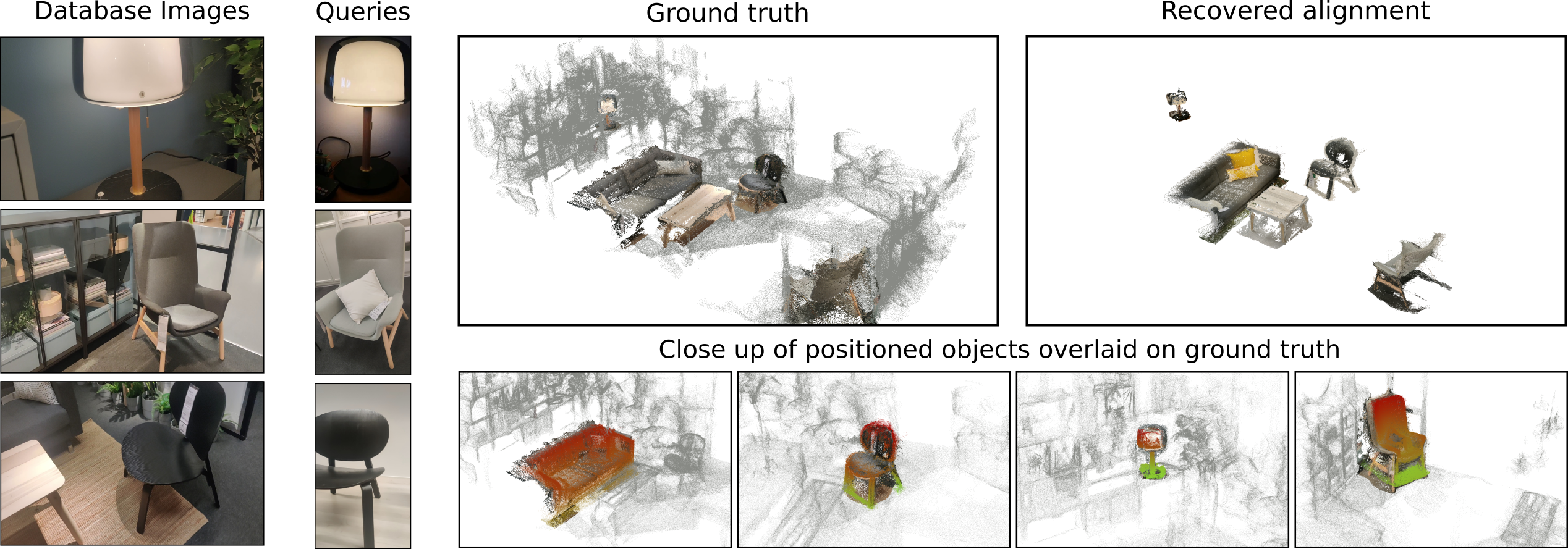}
    \caption{In the context of privacy-preserving localization, we show that it is possible to learn about the content of a scene using camera poses returned by a localization service, without any direct access to the scene representation. (\textbf{1st column}) Examples of images from the scene, used to build the scene representation. The images are shown for illustrative purposes and are not available to an attacker trying to learn about the scene. (\textbf{2nd column}) The attacker queries the service with images of objects, \eg, downloaded from the Internet. 
    (\textbf{3rd \& 4th column}) Using the camera poses for the query image returned by the localization service, the attacker is able to identify the types of objects present in the scene and to accurately place them in the scene. We show the estimated object poses overlaid over the ground truth structure of the scene (which is not accessible to the attacker). 
    The attacker is able to faithfully recover the placement of objects. Overall, our results demonstrate that simple feedback such as camera poses is already sufficient to potentially reveal private details.} 
    \label{fig:teaser}
\end{figure*}

A popular approach for Augmented/Mixed/Virtual Reality (XR) applications is to use a client-server mechanism for localization: 
the user device (client) sends image data 
to a cloud-based system (server) that computes and returns the camera pose~\cite{Middelberg2014ECCV,Ventura2014TVCG,Lynen2020IJRR}. 
Examples of such services include Google's Visual Positioning System \cite{GoogleVPS}, Microsoft's Azure Spatial Anchors \cite{MicrosoftSpatialAnchors}, and Niantic's Lightship \cite{niantic_lightship}. 
Cloud-based localization services are popular for multiple reasons - \textit{first}, performing localization on the server reduces storage requirements 
and 
the computational load, and thus energy consumption, which is important for client devices such as mobile phones and headsets; 
\textit{second}, it enables using robust mapping and localization algorithms that are too expensive for mobile devices; 
\textit{third}, in the context of collaborative mapping, \eg, for the AR cloud or autonomous driving, maintaining a single scene representation in a centralized place is far easier than keeping multiple copies on various mobile devices up-to-date. 

Naturally, sending user data to a server, \eg, in the form of images to be localized or 3D maps recorded by users that will be used for localization, raises privacy concerns~\cite{speciale2019CVPR,Speciale2019ICCV,Dangwal2021ReverseEngineeringPP}. 
Work on privacy-preserving localization aims to resolve these concerns by ensuring that private details cannot be recovered from the data sent ~\cite{Speciale2019ICCV,Dusmanu2021AdversarialSubspacePP,Ng2021NinjaDescPP} to or stored on the server~\cite{speciale2019CVPR,Shibuya2020ECCV,pittaluga2019revealing,Zhou2022ECCV,Do2022CVPR,Geppert2022CVPR,Do2022CVPR}. 

Existing work focuses on scenarios where an attacker gains access to the localization service or can eavesdrop on the communication between client and server. 
In this work, we demonstrate that it is possible for an attacker to learn about the content of a scene stored on a localization server without direct access to the server. 
We show that a localization service will reveal scene-related information through estimated camera poses, \ie, through its normal operation process. 
The attack is based on two recent developments: 
(1) modern visual localization algorithms are designed to be robust against changes such as illumination and seasonal variations~\cite{Toft2021PAMI}. 
This is an essential property for cloud-based localization services in order to operate robustly and reliably. 
However, since these algorithms are robust to (slight) variations in appearance and geometry, they will also localize images showing objects that are similar (but not necessarily identical) to those objects present in the scene. 
(2) massive amounts of images depicting objects in different variations are readily available on the Internet. 
Taken together, both developments allow an attacker to repeatedly query the server with images and to recover the positions of the objects in the scene based on the camera poses returned by the server (\cf Fig.~\ref{fig:teaser}). 
In this paper, we demonstrate the feasibility of this attack by developing a proof-of-concept implementation of the attack.
In summary, we make the following contributions:
\textbf {(1)} we identify a new line of attack in the context of privacy-preserving visual localization based on the camera poses returned by a cloud-based server. 
\textbf{(2)} we show the feasibility of the attack through 
a proof-of-concept implementation of the attack. 
Through experiments, we explore the performance of our implementation as well as the trade-off between localization robustness and potential defenses against the attack. 
\textbf{(3)} the attack is agnostic to the underlying localization algorithm and thus applicable even if the localization system is otherwise perfectly privacy-preserving. 
This paper thus proposes a new research direction for privacy-preserving localization, where the aim for the localization service is to correctly identify whether a query image was taken in the concerned scene or not, in order to prevent leaking information through camera poses.

\section{Related Work}
\textbf{Visual localization.} Most state-of-the-art visual localization algorithms are based on establishing 2D-3D matches between a query image and a 3D model of the scene. 
These correspondences are then used for camera pose estimation. 
The 3D model can either be stored explicitly~\cite{Sattler2017PAMI,sarlin2019coarse,sarlin20superglue,Irschara09CVPR,Li2012ECCV,sun2021loftr,HumenbergerX20Kapture,Panek2022ECCV}, \eg, in the form of a Structure-from-Motion (SfM) point cloud, or implicitly in the form of the weights of a  machine learning model~\cite{brachmann2020ARXIV,Brachmann2019ICCVa,Brachmann2018CVPR,Shotton2013CVPR,Valentin20163DV,Cavallari20193DV}. 
In the former case, local feature descriptors are associated with 3D points of the model. 
It has been shown that this information is sufficient to recover detailed images from the 3D map~\cite{pittaluga2019revealing,Song2020ECCV}, although sparsifying these models~\cite{Camposeco2019CVPR,Yang2022CVPR} might effectively make them privacy-preserving~\cite{Chelani2021CVPR}. 
Approaches based on implicit representations map image pixels or patches to 3D points by training scene coordinate regression models~\cite{Shotton2013CVPR,brachmann2020ARXIV}. 
Recently, it was claimed that such approaches are inherently privacy-preserving~\cite{Do2022CVPR}. 
However, feature-based methods currently scale better to large scenes and are able to better handle condition changes~\cite{Toft2021PAMI}, such as illumination or seasonal changes, between the query image and the database images used to build the the scene representation. 
The resulting robustness is highly important in many applications of visual localization, including AR and robotics. 
The robustness is a direct consequence of recent advances in  local features~\cite{detone2017superpoint,Dusmanu2019CVPR,r2d2} and effective feature matchers~\cite{sarlin20superglue,Zhou2021CVPR,sun2021loftr,wang2022Matchformer}. 
In this paper, we show that robustness to changing conditions enables an attacker to learn about the content of the scene: 
robustness to changing conditions not only bridges the gap between (small) variations in scene appearance and geometry observed in images depicting the same place, but also leads to correspondences between images depicting similar but not identical objects, \eg, different chairs. 
In turn, these correspondences can be used to localize the object in the scene, which is the basis for the attack described in this work. 
Note that the properties we exploit are inherent to robust localization algorithms and are not restricted to feature-based methods. 
Ultimately, any robust localization system needs to handle variations in shape and appearance. 

\PAR{Privacy-preserving visual localization.} 
Existing work on privacy-preserving localization focuses on two points of attack: 
(1) ensuring that data sent to a localization service does not reveal private information. 
(2) ensuring that data stored on a localization service does not reveal private information. 
For the former case, it has been shown that images can be recovered from local features~\cite{Weinzaepfel2011CVPR,Dosovitskiy2016CVPR,Dangwal2021ReverseEngineeringPP}. Work on privacy-preserving queries to a localization server thus mostly aims at developing features that prevent image recovery~\cite{Dusmanu2021AdversarialSubspacePP,Ng2021NinjaDescPP} or on obfuscating the feature geometry~\cite{Speciale2019ICCV,Geppert2020ECCV}. 
Similarly, work on privacy-preserving scene representation aims to obfuscate the geometry~\cite{speciale2019CVPR,shibuya2020privacy} (although scene geometry can be recovered under certain conditions~\cite{Chelani2021CVPR}), splitting the maps over multiple server for increased data security~\cite{Geppert2022CVPR}, using implicit representations~\cite{Do2022CVPR}, or storing raw geometry without any feature descriptors~\cite{Zhou2022ECCV}. 

This paper presents a new line of attack that complements existing work. 
Previous work considers a scenario where the attacker gains access to the service. 
In contrast, we show that it is possible to recover scene content from the very basic information provided by any localization service, namely the camera poses estimated for query images. 
As such, the attack is still feasible even if the data send to and stored on the server is completely privacy-preserving. 
Our work thus shows that existing privacy-preserving localization approaches are not sufficient to ensure user privacy.

\section{Recovering Scenes from Camera Poses}\label{sec:attack}
Any localization system returns the camera poses of localized query images. 
At the same time, modern localization algorithms aim to be robust to shape and appearance variations in order to be robust to changes in viewing conditions. 
This feature, however, opens up the possibility that not only genuine queries, but also images of objects that are similar to the ones present in the scene can be localized. 
The camera poses of the localized images can then in turn be used to infer the positions of certain objects in the scene, potentially revealing more information about the scene than the cloud-based service / a user would like to disclose.

 Naturally, an attacker does not know which objects are present in the scene and thus which images to use for their queries. 
 The Internet is a source of a theoretically 
unlimited number of images, videos, and 3D models of objects of different types and appearances. 
This naturally leads to an idea of a potential attack, where an attacker just downloads such images and videos, bombards the server with localization requests, and uses poses of localized images to reveal detailed scene structure. 

In the following sections, we investigate this new type
of attack, and we try to answer several questions: Can an attacker with access 
to images 
and videos of objects similar to those present in the scene 
easily learn about the presence/absence of different objects and their placement in the scene just from the poses returned by a localization service? What are the challenges of such an
attack, and are these challenges easily solvable? Can cloud-based services easily prevent such attacks? 
To this end, we present a proof-of-concept implementation of the attack.\footnote{We only aim to show feasibility. We believe that better attack algorithms are certainly possible.} 
Later, Sec.~\ref{sec:discussion} discusses an approach to potentially mitigate the attack and why its 
effectiveness is limited.


\subsection{Formalization} \label{sec:attack_outline}

We assume a localization server $\mathcal{L}$ that is responsible for localizing images in a scene $\mathcal{S}$. $\mathcal{L}$  tries to align each query image it receives with the scene representation as best as possible. 
If an image can be localized, the server returns a 6-dof camera pose $[\mathbf{R}|\mathbf{t}]$. 
We assume that the scale of the translation component $\mathbf{t}$ is in known. 


An adversary $\mathcal{A}$ is querying $\mathcal{L}$ with many images of different objects, where each image contains only one dominant object to avoid confusion about which object from the image was localized in the scene. 
$\mathcal{A}$, using the poses returned by $\mathcal{L}$, wants to learn about the presence/absence of objects in the scene $\mathcal{S}$, 
and wants to infer their (approximate) positions. 
As such, $\mathcal{A}$ tries to construct an (approximate) ”copy” of the scene $\mathcal{S}$ or at least its layout. 

In this setting $\mathcal{A}$ needs to deal with two 
challenges:
\begin{enumerate}
    \item $\mathcal{A}$ queries $\mathcal{L}$ with images of objects that, in general, differ geometrically from the actual objects in the scene. 
    In the best case, the pose returned by the server provides the best-possible approximate alignment between the query and actual object. In general, the returned poses will be noisy and can be quite inaccurate if only a part of the object, \eg, a chair's leg, is aligned. 
    Creating an accurate "copy" of the scene from such poses is a challenging problem. 

    \item $\mathcal{A}$ has, in general, no a-priori information about the type of the scene and which objects are visible in it. 
    Since $\mathcal{L}$ can also return poses for objects that are not in the scene, $\mathcal{A}$ needs to have a mechanism for deciding the presence/absence of an object based on the returned poses. Naturally, having to deal with noisy and inaccurate poses makes the decision process harder.
\end{enumerate}
In general, it is not possible to overcome these challenges by using a single image of each object. A single camera pose returned by $\mathcal{L}$, without additional information, does not provide enough data for deciding about the presence/absence of the object in the scene and the quality of the pose. 

However, given the large amount of images available on the Internet, and in particular the availability of videos, 
$\mathcal{A}$ can use several images of the same object taken from different viewpoints. 
Jointly reasoning about all of the corresponding poses obtained for  these images can then be used to decide the presence and position of the object. 


\begin{figure*}[t!]
    \centering
    \includegraphics[width = 0.95\textwidth]{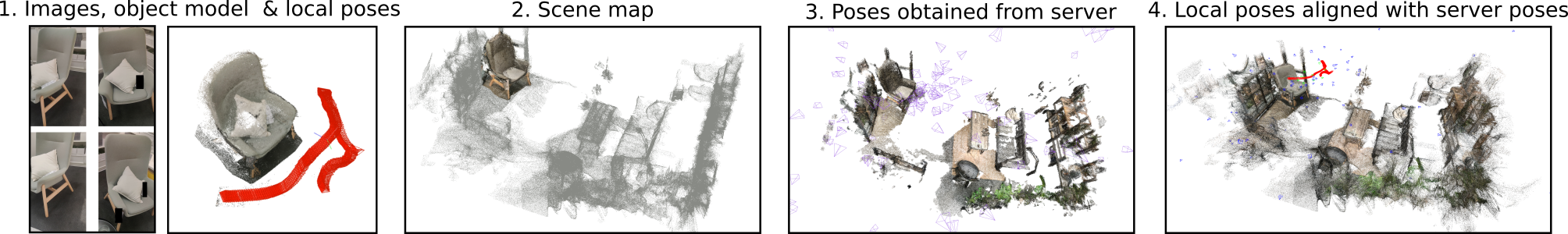}
    \caption{\textbf{Object alignment example}: \textbf{1.} A 3D model $\mathcal{M}$ of an object and corresponding camera poses $\mathbf{P}_{o}$ in the attacker's local coordinate system, built from a sequence of object images. \textbf{2.} The server scene with a similar object. \textbf{3.} The noisy poses returned by the server for the queried object images. \textbf{4.} Sequences of local and server-provided poses aligned to approximately place the object in the scene.}
    \label{fig:aligment_example}
\end{figure*}

\subsection{3D Object Placement}\label{sec:postioning}
Assuming that the attacker knows that an object is present, they still need to predict its position and orientation in the scene based on the pose estimates provided by the server. 
To this end, the attacker can use that multiple images of the same object taken from different viewpoints are available. 
%
%
%
These images can be used by $\mathcal{A}$ to 
build a local 3D model $\mathcal{M}$, \eg, using SfM~\cite{Schoenberger2016CVPR} and MVS~\cite{Schoenberger2016ECCV}, and to compute the poses $\mathbf{P}_{o}$ of these images \wrt this model. 
In turn, $\mathcal{L}$ provides a set $\hat{\mathbf{P}}_{o}$ of poses for (a subset of) these images in the coordinate system of the scene model $\mathcal{S}$. 
The problem of placing the object in the copy of the scene $\mathcal{S}$ thus reduces to the problem of aligning both sets of poses (\cf Fig.~\ref{fig:aligment_example}). 
%
The camera poses $\hat{\mathbf{P}}_{o}$ provided by $\mathcal{L}$ can be very noisy and can contain outliers. 
Thus, the alignment process needs to be robust.

\begin{algorithm}[t!]
    \caption{Best single camera based alignment between sets of poses}
    \label{Alg:Alignemnt} 
    \small
    \hspace*{\algorithmicindent} \textbf{Input} $\mathbf{P}_{o} = \{[\mathbf{R}_i| \mathbf{t}_i]\}$,$\mathbf{\hat{P}}_{o} = \{[\mathbf{\hat{R}}_i| \mathbf{\hat{t}}_i]\}$, $\delta_{r}$,$\delta_{t}$ \\
    \hspace*{\algorithmicindent} \textbf{Output} $\textbf{R}_{best}, \textbf{t}_{best}, \epsilon$
    \begin{algorithmic}[1]
    \Procedure{Get-Best-Alignment}{}
    \State $\texttt{N} \gets {|\mathbf{P}_{o}|}$
    \State $\texttt{Inliers\_best} \gets \phi$
    \For{\texttt{i = 1 to N}}
        \State $\mathbf{R}_{est} \gets \mathbf{\hat{R}}^{\top}_i \mathbf{R}_i$
        \State $\mathbf{t}_{est} \gets \mathbf{\hat{R}}^{\top}_i(\mathbf{t}_i - \mathbf{\hat{t}}_i )$
        \State $\texttt{Inliers} \gets \phi$
        \For{\texttt{j = 1 to N}}
            \State $\Delta_{r} \gets \angle(\mathbf{R}_j \mathbf{R}^{\top}_{est} \mathbf{\hat{R}}_j^{\top})  $
            \State $\Delta_{t} \gets ||\mathbf{\hat{R}}^{\top}_{j} \mathbf{\hat{t}}_j - \mathbf{R}_{est} \mathbf{R}^{\top}_{j} \mathbf{{t}}_j + \mathbf{t}_{est})||$
            \If{$\Delta_{r} < \delta_{r}$ and $\Delta_{t} < \delta_{t}$}
                \State $\texttt{Inliers} \gets \texttt{Inliers} \cup \{\texttt{j}\}$
            \EndIf
        \EndFor
        \If{ $|\texttt{Inliers}| > |\texttt{Inliers\_best}|$}
             \State $\texttt{Inliers\_best} \gets \texttt{Inliers}$
        \EndIf
    \EndFor
    \State $\epsilon \gets |\texttt{Inliers\_best}|/\texttt{N}$
    \State $\mathbf{R}_{best}, \mathbf{t}_{best} \gets$ \textbf{Average(\texttt{Inliers\_best})}
    \EndProcedure 
    \end{algorithmic}
\end{algorithm}

As mentioned above, for simplicity we assume that the scale of the 3D model stored by $\mathcal{L}$ is known.\footnote{In the context of user-generated maps, captured by devices with IMUs such as mobile phones or dedicated XR headsets, it seems realistic to assume that the scale of the maps is provided in meters.} 
Similarly, the scale of the local model $\mathcal{M}$ can be (approximately) recovered using the known size of the object. 
In this case, the two poses, in the coordinate systems of $\mathcal{M}$ and $\mathcal{S}$, for a single image already provide an alignment hypothesis, \ie, the relative pose between them. 
As outlined in Alg.~\ref{Alg:Alignemnt}, we evaluate all hypotheses. 
The input to Alg.~\ref{Alg:Alignemnt} are the two sets of poses, $\mathbf{P}_{o}$ and 
$\hat{\mathbf{P}}_{o}$, 
and two error thresholds - $\delta_{r}$ for rotation and $\delta_{t}$ for translation. 
 For each pair of corresponding camera poses - local and server-provided, a relative transformation is computed (line 5-6). 
 One set 
 of poses is transformed using this estimated transformation, and errors for rotation and translation between corresponding pairs are computed (Lines 9-10). Using the two thresholds, we determine which other pose pairs are inliers to the pose hypothesis 
 (Lines 11-12). 
 The transformation with the largest number of inliers is selected (Lines 13-14) and refined by averaging the relative poses of all inliers. 
 
 Obviously, not knowing the scale of the scene model $\mathcal{S}$ is insufficient to prevent the attacker from placing the object in the scene as the scale and relative transformation can be recovered from two pairs of poses. 
 Additionally, there are ways to further robustify the alignment process. 
 \Eg, if images of multiple very similar instance of an object and the corresponding 3D models are available, it seems reasonable to assume that images of different instances taken from similar viewpoints will also result in similar pose estimates by $\mathcal{L}$. 
 These estimates can then be used to average out noise in the poses. 
 Similarly, the relation between different objects, \eg, a monitor standing on a desk, can be used to stabilize the process of placing objects in the scene. 
 However, we do not investigate such advanced strategies in this paper. 
 
 

\subsection{Deciding the Presence/Absence of an Object}\label{sec:deciding}
We assume that  $\mathcal{L}$ is running a localization algorithm that is robust to shape and appearance variations and that is aligning query images to the scene as best as it can. 
At the same time, $\mathcal{L}$ can also return poses for objects that are not in the scene, as well poses for objects that are not even from the same categories or similar to objects present in the scene. Deciding if an object is present or not in a scene based on the poses returned for its images by the localization server is therefore a challenging problem.

\begin{figure*}[t]
    \centering
    \includegraphics[width=0.90\linewidth]{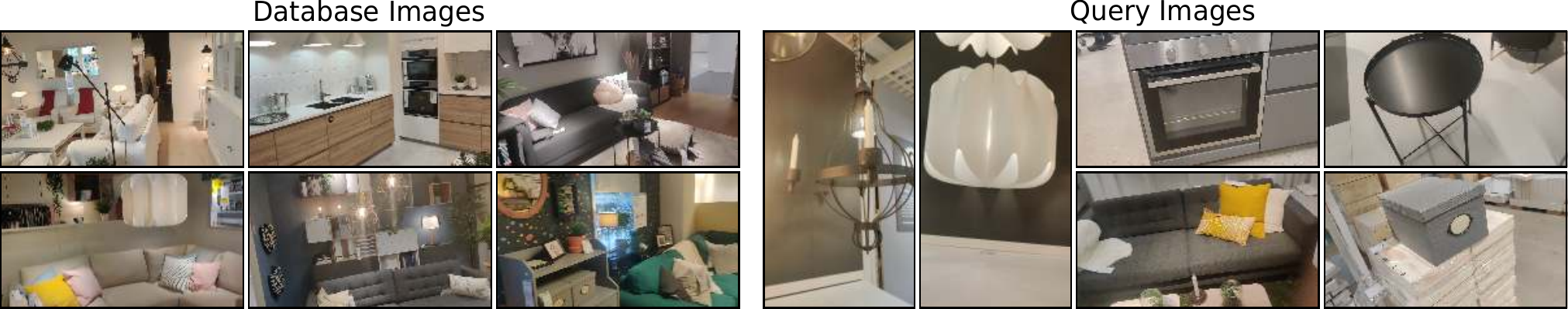}
    \caption{Example images from \textit{IKEA-Scenes} (left) and one of the objects of corresponding scenes in \textit{IKEA-Objects} (right).}
    \label{fig:ikea_examples}
\end{figure*}

For an attacker $\mathcal{A}$ trying to recover scene information via camera poses, it is impossible to determine which type of objects are present 
using just 
a single camera pose returned for one query image of each of the objects. 

To overcome this challenge, $\mathcal{A}$ can employ several possible techniques; \eg, 
they can use statistics about object co-occurrence to select the set of queries and associated camera poses having a high probability of their spatial distribution.  
Another simple solution is to use multiple images of the same object taken from different viewpoints or to cluster query images into groups depicting similar objects that are assumed to be matched with the same object in the scene $\mathcal{S}$.
$\mathcal{A}$ can then use different images from these groups to query $\mathcal{L}$ and decide on the presence/absence of the object based on the consistency of returned poses.
Even though the returned poses can be noisy and can contain outlier poses, in general, it is expected that a reasonably large subset of images depicting the same object from different viewpoints or depicting objects from the same group will show consistency of returned poses if a similar object is present in $\mathcal{S}$.  
On the other hand, poses obtained for images 
of an object that is absent can be expected to exhibit a much higher variance.

In this paper, we discuss and evaluate 
another strategy for presence/absence decision that allows us to show the completeness of the attack and present its proof-of-concept implementation.
We assume that the attacker $\mathcal{A}$ learns certain statistics for each object or category from a curated training data that comprises of scenes with known presence/absence of these objects or object categories.
This can be done for different types of localization schemes over huge amounts of 3D data. 
The attacker can then use these learned statistics to infer the presence/absence of objects when attacking an unknown scene $\mathcal{S}$. 

For experimental results in the later sections, we use the inlier-ratio $\mathbf{\epsilon}$ 
obtained from the object positioning step (Line 15 in Alg.~\ref{Alg:Alignemnt}) as this statistic. 
We can assume that for each object (or 
a class of objects) $o$,
$\mathcal{A}$ has inlier-ratios $\mathbf{\epsilon}^{+}_{o}$ and $\mathbf{\epsilon}^{-}_{o}$ that are trained on scenes with known presence or absence of $o$. \Eg, $\mathbf{\epsilon}^{+}_{o}$ and $\mathbf{\epsilon}^{-}_{o}$ can be computed as the medians of $\mathbf{\epsilon}_{o}$ over all "present(+)/absent(-)" scenes. 
Based on these statistics, the presence or absence of $o$ in the unknown scene $\mathcal{S}$ can be decided by comparing the distances of $\mathbf{\epsilon}^{\mathcal{S}}_{o}$ to $\mathbf{\epsilon}^{+}_{o}$ and $\mathbf{\epsilon}^{-}_{o}$.
%
We provide concreteness to this idea when assessing its effectiveness over a real world dataset in Section \ref{sec:classify_eval_ikea}. 

\begin{figure*}[t]
    \centering
    \includegraphics[width = 0.9\linewidth]{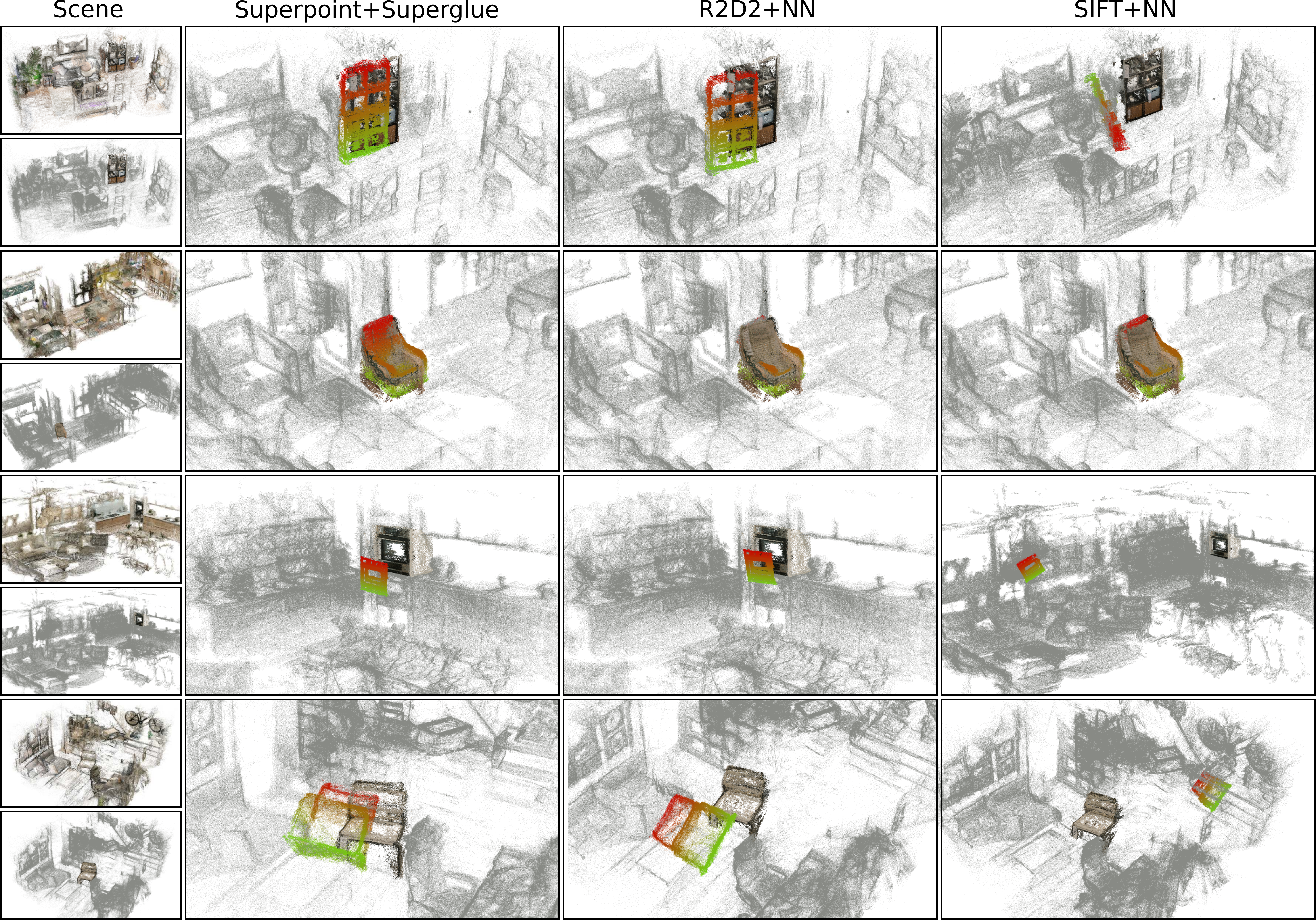}
    \caption{Qualitative results for aligning objects in different scenes of the \textit{IKEA-Scenes} dataset. We evaluate three combinations of local features and matchers. Aligned objects are color-coded green to red along the gravity direction to make their orientation better visible.} 
    \label{fig:qualitative_position}
\end{figure*}

\section{Datasets}
We use multiple 
datasets for our experiments:
    
    \noindent \textbf{IKEA-Scenes} and \textbf{IKEA-Objects} - We captured image sequences of seven different inspiration-rooms at an IKEA furniture store (\cf Fig.\ref{fig:ikea_examples}). 1,000-2,500 images were captured for each room, 
    depending on its size. 4-10 objects from each room were selected, and a separate sequence of images was captured for each of them in the inventory section of the store, where the surrounding environment was different from that of the inspiration rooms. Note that the two instances of each object have the same model, but in many cases 
    differ in color and size. Presence/absence of additional objects such as cushions on a sofa, or a computer on a desk can additionally change the overall appearance of the two instances. In total, the dataset comprises 38 object instances covered by 100-200 images each. 
    While capturing the dataset, we 
    tried to only have a single object occupying a large part of each image. 
    However, this was not always possible and no post processing has been applied to mask out objects. We call the inspiration-room data 
    \textit{IKEA-Scenes} and the data from the inventory section 
    \textit{IKEA-Objects}.   
    
    \noindent \textbf{ScanNet-Office-Scene} - To show that the objects do not need to be of the exact same model for the proposed attack to work, we consider a generic office scene - \textit{scene0040} from the ScanNet \cite{dai2017scannet} dataset. 
    
    \noindent \textbf{Office-Objects} - We collected image sequences of 5 common office room objects - a \textit{door}, a \textit{whiteboard}, an \textit{office chair}, a \textit{desk with computer}, and a \textit{bookshelf}. These images are used as queries by the attacker.
    
    \noindent \textbf{RIO10} - RIO10 \cite{wald2020} is a 
    localization benchmark 
    dataset which we use to evaluate the effectiveness of a potential 
    defence strategy that a localization server might employ.

We manually scale all local 3D models constructed by the attacker to  roughly metric scale. 
\vspace{-0.1cm}
\section{Experimental Evaluation} 
\label{sec:experiments}

This section presents a series of experiments that show the practical feasibility of the attack introduced in Sec.~\ref{sec:attack}. 
First, we show via qualitative results that the method proposed in Sec.~\ref{sec:postioning} allows the attacker to place the 3D models of 
relevant objects close to the actual corresponding objects in the scene.  
We then explain and evaluate a simple implementation of the method described in Sec.~\ref{sec:deciding} that the attacker can use to decide the presence/absence of objects. 

For querying the localization server, we use images from the datasets described above. 
To implement the server, we use HLoc~\cite{sarlin2019coarse,sarlin20superglue} (with default thresholds and parameters), a state-of-the-art visual localization approach. 
HLoc uses feature descriptors to establish 2D-3D matches between features extracted from the query image and 3D scene points. 
The resulting correspondences are then used for pose estimation. 
We demonstrate the reliance of the attack on the robustness of the localization process by evaluating 
three different local image features and matchers: Superpoint~\cite{detone2017superpoint} features with the SuperGlue\cite{sarlin20superglue} (most robust), R2D2 \cite{r2d2} with  Nearest Neighbor (NN) matching, and  SIFT \cite{Lowe04IJCV} with NN matching (least robust). 

\subsection{3D Object Placement}\label{sec:exp_position}


\begin{figure*}
    \centering
    \includegraphics[width = 0.88\linewidth]{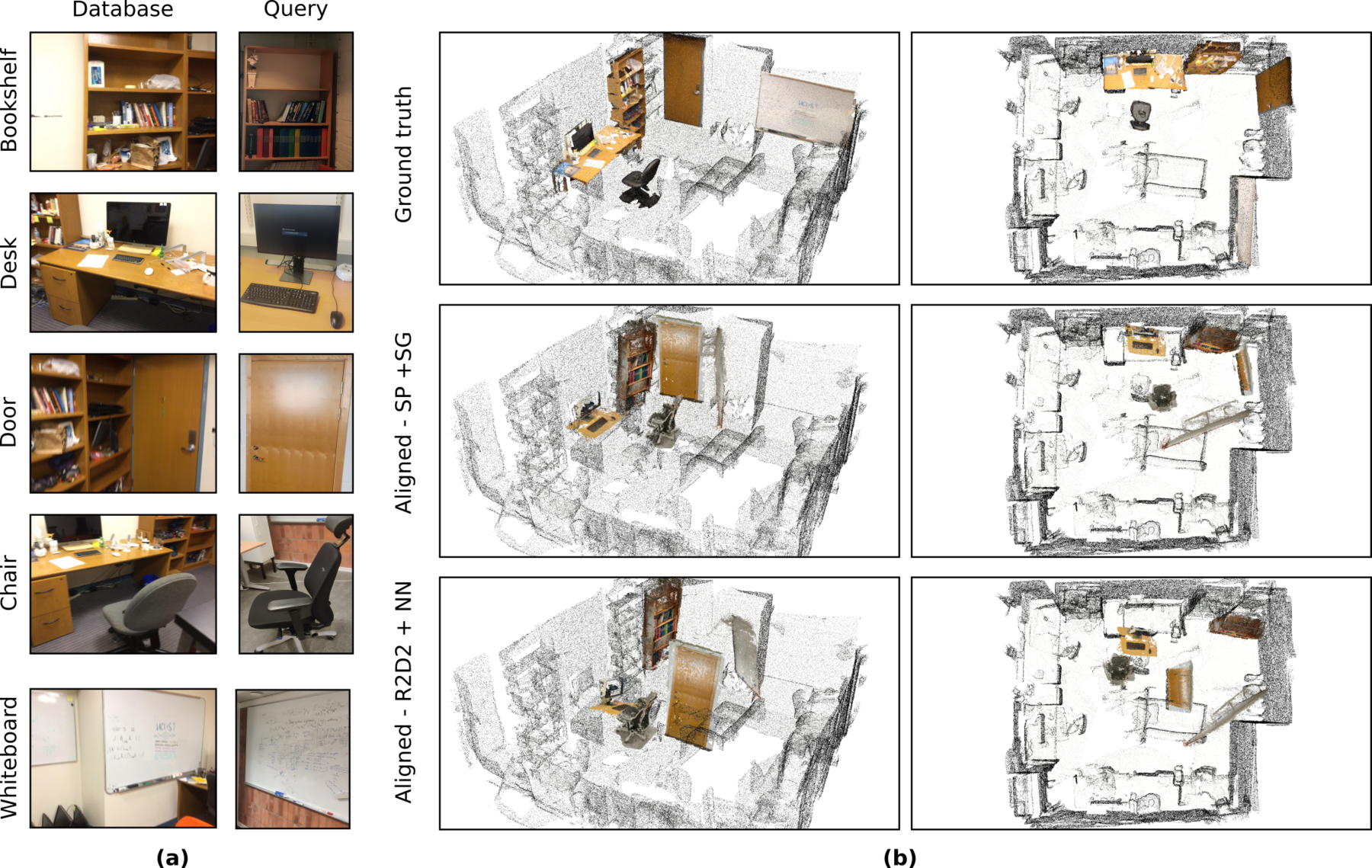}
    \caption{\textbf{(a)} Example images from \textit{ScanNet-Office-Scene} and corresponding objects in \textit{Office-Objects}. \textbf{(b)} Qualitative results for aligning generic office objects in ScanNet \cite{dai2017scannet} \textit{scene0040}, using Superpoint+Superglue and R2D2+NN.}
    \label{fig:qualitative_position_Scannet}
\end{figure*}

We qualitatively evaluate the accuracy of the 3D object placements obtained using the approach from Sec.~\ref{sec:postioning} for the \textit{IKEA-Scenes} and \textit{ScanNet-Office-Scene} datasets. 
We use qualitative results rather than quantitative metrics since it is hard to quantify when a placement is realistic enough. 
\Eg, consider the predicted positions of the oven in the 3rd row of Fig.~\ref{fig:qualitative_position}. 
The first two predictions are far enough from the ground truth position that a metric such as the IoU of the 3D bounding boxes of the objects will discard them as wrong. 
Yet, the estimated positions are close enough to the ground truth to provide the attacker with a good layout of the scene.  

Fig.~\ref{fig:qualitative_position} shows results for  placing selected items from the \textit{IKEA-Objects} dataset in 4 different scenes from the \textit{IKEA-Scenes} dataset. 
Fig.~\ref{fig:qualitative_position_Scannet} shows results for placing objects from the \textit{Office-Objects} dataset in the \textit{ScanNet-Office-Scene} dataset. 
As can be seen, using a robust localization process based on Superpoint features and the Superglue matcher or R2D2 features allows the attacker to place the objects close to their ground truth positions. 
In particular, the results from Fig.~\ref{fig:qualitative_position_Scannet} show that the alignment also works well when the queried object is not the same model of different color/size but also a very different one in terms of shape and overall appearance. 
The results clearly demonstrate 
the practical feasibility of the placement strategy. 

We used slightly different values for the error thresholds required by the positioning algorithm based on the object size and obtained poses. 
Such an approach is feasible if a human supervises the attack. Code and data is available at \href{https://github.com/kunalchelani/ObjectPositioningFromPoses}{\small{https://github.com/kunalchelani/ObjectPositioningFromPoses}}. 

\subsection{Deciding the Presence/Absence of an Object}{\label{sec:classify_eval_ikea}}

\begin{table*}[t!]
\centering
\scriptsize
\begin{tabular}{|c||c|c|c|c|c|c||c|c|c|c|c|c||c|c|c|c|c|c|}
\hline
\multirow{3}{*}{Scene} & \multicolumn{6}{c||}{Superpoint+Superglue}                                                                                      & \multicolumn{6}{c||}{R2D2+NN}                                                                                                      & \multicolumn{6}{c|}{SIFT + NN}                                                                                                  \\ 
\cline{2-19}
                       & \multicolumn{2}{c|}{$10^{\circ},0.25m $} & \multicolumn{2}{c|}{$30^{\circ}, 0.5m $} & \multicolumn{2}{c||}{~ $60^{\circ}, 2m $} & \multicolumn{2}{c|}{$10^{\circ}, 0.25m $} & \multicolumn{2}{c|}{$30^{\circ}, 0.5m $} & \multicolumn{2}{c||}{~ $60^{\circ}, 2m $} & \multicolumn{2}{c|}{$ 10^{\circ}, 0.25m $} & \multicolumn{2}{c|}{$30^{\circ}, 0.5m $} & \multicolumn{2}{c|}{$ 60^{\circ},2m $}  \\ 
\cline{2-19}
                       & Precision & Recall                       & P    & R                                 & P    & R                                  & P    & R                                  & P    & R                                   & P    & R                                  & P    & R                                   & P    & R                                 & P    & R                                \\ 
\hline
Scene1                 & 0.6       & 0.85                         & 0.75 & 0.85                              & 0.67 & 0.85                               & 0.57 & 0.57                               & 0.36 & 0.57                                & 0.28 & 0.57                               & 0.33 & 0.57                                & 0.45 & 0.71                              & 0.33 & 0.43                             \\
Scene2                 & 0.36      & 0.4                          & 0.36 & 0.5                               & 0.37 & 0.6                                & 0.34 & 0.4                                & 0.3  & 0.3                                 & 0.35 & 0.6                                & 0.33 & 0.4                                 & 0.26 & 0.5                               & 0.28 & 0.6                              \\
Scene3                 & 0.55      & 0.71                         & 0.36 & 0.57                              & 0.25 & 0.43                               & 0.31 & 0.71                               & 0.47 & 1                                   & 0.41 & 1.0                                & 0.3  & 0.42                                & 0.5  & 0.42                              & 0.44 & 1.0                              \\
Scene4                 & 0.17      & 0.4                          & 0.23 & 0.6                               & 0.14 & 0.4                                & 0.34 & 0.6                                & 0.28 & 0.4                                 & 0.2  & 0.4                                & 0.15 & 0.4                                 & 0.15 & 0.4                               & 0.17 & 0.4                              \\
Scene5                 & 0.33      & 0.6                          & 0.4  & 0.8                               & 0.44 & 0.8                                & 0.5  & 0.6                                & 0.34 & 0.4                                 & 0.5  & 0.6                                & 0.22 & 0.4                                 & 0.25 & 0.4                               & 0.33 & 0.4                              \\
Scene6                 & 0.25      & 0.6                          & 0.28 & 0.6                               & 0.22 & 0.4                                & 0.22 & 0.4                                & 0.3  & 0.6                                 & 0.33 & 0.8                                & 0.14 & 0.2                                 & 0.2  & 0.2                               & 0.25 & 0.6                              \\
Scene7                 & 0.5       & 0.5                          & 0.5  & 0.33                              & 0.33 & 0.5                                & 0.6  & 0.5                                & 0.5  & 0.5                                 & 0.38 & 0.5                                & 0    & 0                                   & 0.14 & 0.17                              & 0    & 0                                \\
\hline
\end{tabular}
\caption{{Precision (P)} and {recall (R)} 
of our method to determine the presence of objects for the \textit{IKEA-scenes} and \textit{IKEA-Objects} datasets.}
\label{tab:pr}
\end{table*}

\begin{figure*}[t]
    \centering
    \begin{subfigure}[b]{0.3\textwidth}  
        \centering 
        \includegraphics[width=\textwidth]{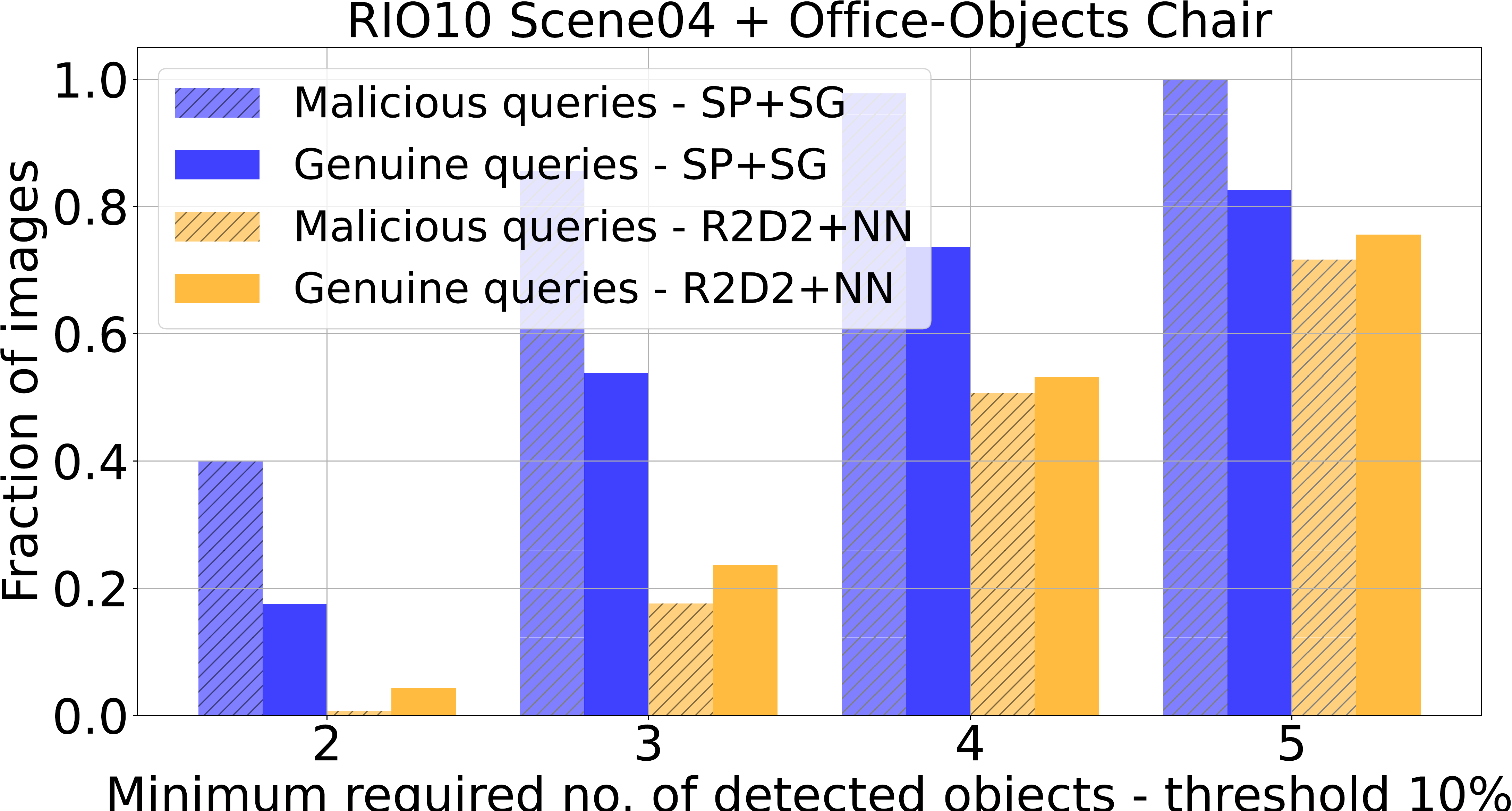}
    \end{subfigure}
    \hspace{5pt}
    \begin{subfigure}[b]{0.3\textwidth}   
        \centering 
        \includegraphics[width=\textwidth]{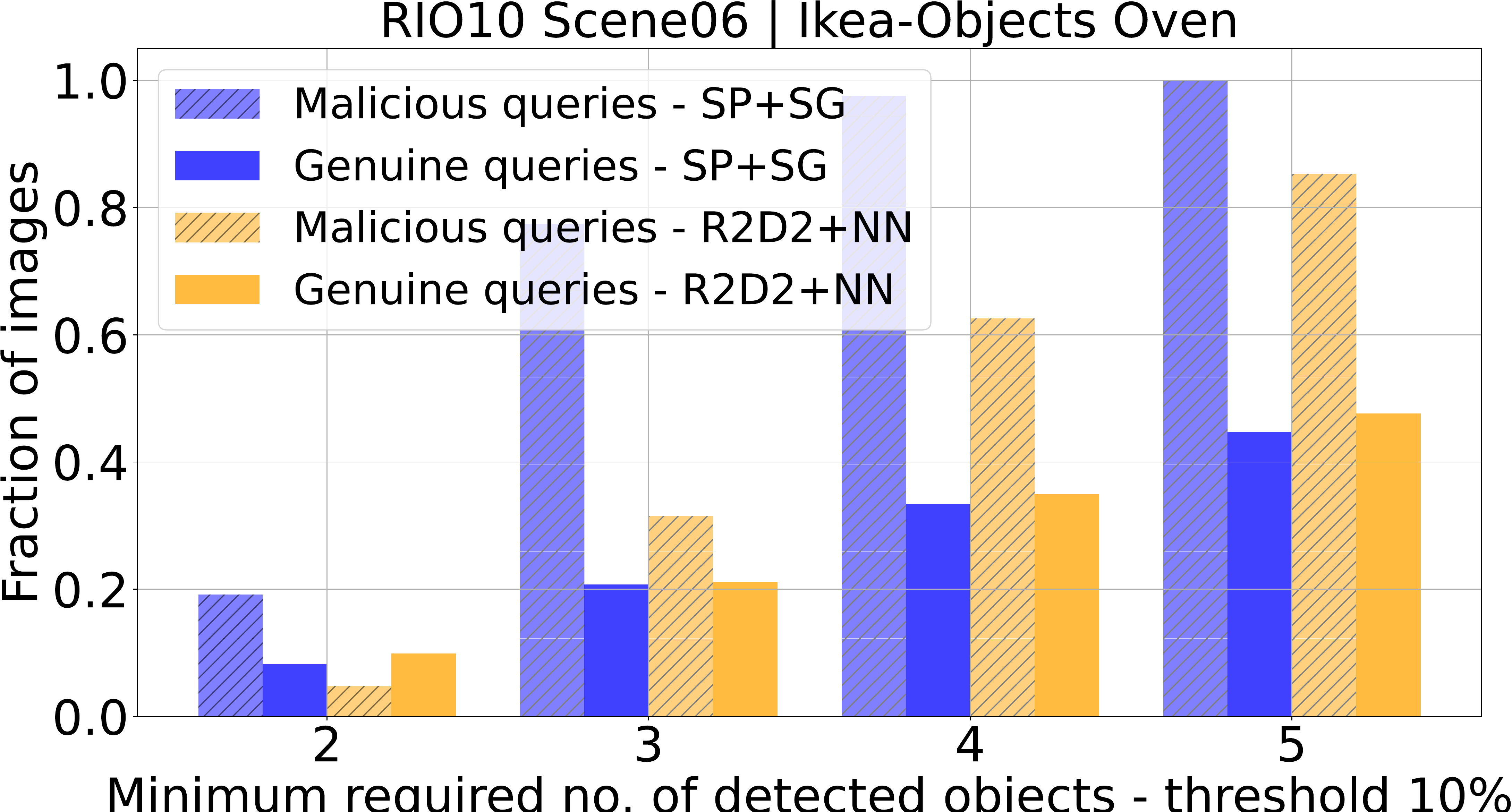}
    \end{subfigure}
    \hspace{5pt}
    \begin{subfigure}[b]{0.3\textwidth}   
        \centering 
        \includegraphics[width=\textwidth]{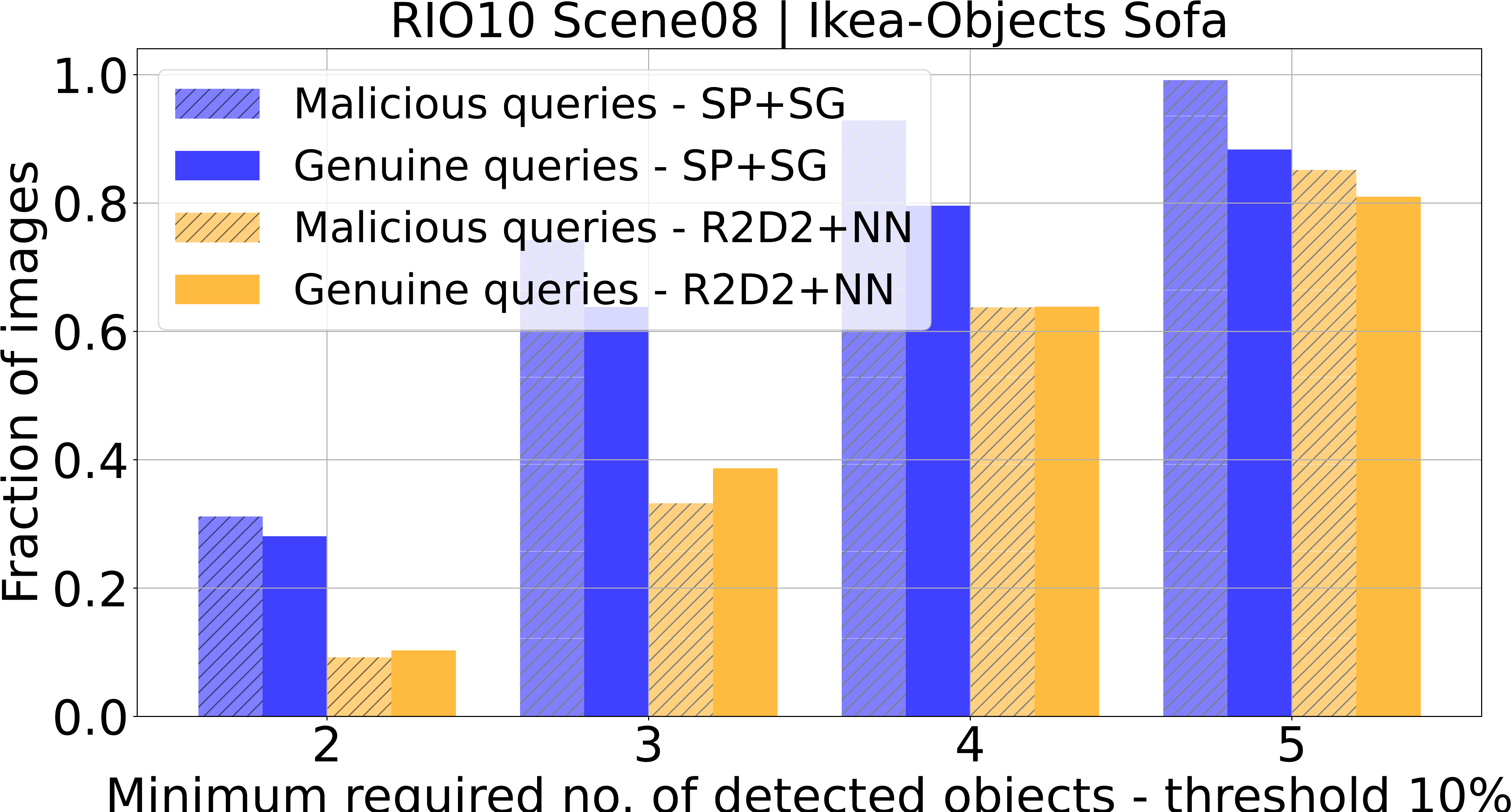}
    \end{subfigure}
    \caption{Effectiveness of a potential approach to prevent the proposed attack based on not providing poses for queries containing only a few objects. 
    Only objects contributing at least 10\% of the inliers found on the object with the most inliers are considered. As can be seen, finding a suitable threshold for the minimum number of visible objects can be difficult.}
    \label{fig:inlier_thresh}
    \vspace*{-0.1cm}
\end{figure*}

In Sec.~\ref{sec:deciding}, we suggested 
strategies which an attacker can use to decide whether 
an object is present or not in a scene $\mathcal{S}$. 
for each object. 

Concretely, using a set of training scenes, the attacker has learned representative values $\epsilon^{+}$ and $\epsilon^{-}$ for the inlier-ratio returned by Alg.~\ref{Alg:Alignemnt} for cases where the object is present(+) respectively absent(-).  
%
When deciding the presence of an object $\mathbf{o}$ in a scene $\mathcal{S}$, the attacker uses the inlier ratio ($\epsilon$) from Alg.~\ref{Alg:Alignemnt} to make their decision. 
The object $\mathbf{o}$ is considered to be present in the scene if $|\epsilon - \epsilon^{+}| < |\epsilon - \epsilon^{-}|$ and otherwise considered as absent.

We use the \textit{IKEA-Scenes} and \textit{IKEA-Objects} dataset for this experiment. When deciding the presence/absence 
of an object in a scene, the other 6 scenes are used as training scenes. 
Many of the objects from \textit{IKEA-Objects} are only present in one of the scenes from \textit{IKEA-Scenes}. 
In these cases, no reference value for $\epsilon^{+}$ is available for these scenes. 
In such cases, the object is considered as present if $\epsilon > \epsilon^{-}$. 
This strategy is motivated by the assumption that correctly placing an object that is present results in a higher inlier-ratio than placing objects that are not present. 

Tab.~\ref{tab:pr} shows precision and recall of this strategy. 
Since the computation of the inlier-ratio $\epsilon$ depends upon the error thresholds, 
we present the results for three different sets of thresholds. 
The results show that for most scenes, it is possible to obtain a precision/recall of approx. 0.4/0.6, which, 
\eg, translates to 3 out of 5 present, and around 29 out of 33 absent objects from \textit{IKEA-objects} being correctly classified. The average precision using random guessing in these scenes is 0.19. 
This, together with the quality of the placement, clearly validates the feasibility of the proposed attack.

\color{black}



\section{Preventing the Attack?}
\label{sec:discussion}
\vspace{-0.1cm}
A natural way to prevent the presented attack is to try to distinguish between genuine and malicious queries. 
By not sending poses for query images deemed as (potentially) malicious, the localization service effectively prevents the attacker from using pose estimates to learn about the scene. 

One potential classification strategy is based on the fact that the attacker sends images focusing on a single object. 
In this case, we expect 
that most of the 3D points from the inlier 2D-3D matches found by HLoc lie on a single 3D object. 
We thus count the number of 3D objects that contribute at least a certain fraction of inliers 
(X\% of the inliers of the object contributing the largest number of inliers). 
If the number is too small, the query image is considered to be malicious and is rejected. 

Fig.~\ref{fig:inlier_thresh} shows results for three different objects used to attack three different scenes of the RIO10 dataset \cite{wald2020}. 
Here, we use the instance-level labels provided by the dataset, which include background classes such as floor and walls, to define objects. 
As can be seen, rejecting the majority of malicious queries while retaining genuine queries can be challenging. 
The reason is that even while focusing on a single object, other objects might be partially visible in the queries, \eg, part of a desk for monitors, different pillows on a couch, books on a shelf, \etc 
In addition, genuine queries might focus on small parts of the scene or even individual objects. 
Thus, finding a suitable threshold on the minimum number of visible objects can be hard. 
Furthermore, note that this defense strategy requires the service to have knowledge about the objects in the scene, either extracted from the queries or the scene representation. 
This requirement creates a potential privacy risk if an attacker is able to gain access to the service. 


\vspace{-0.1cm}
\section{Conclusions and Future work}
\vspace{-0.05cm}
In this paper, we have considered the problem of privacy-preserving localization. 
Prior work aims to defend attacks for the case where the attacker gains access to a cloud-based localization service. 
In contrast, we show that it is possible for an attacker to recover information about the scene by using the service as intended: 
by querying the server with images of different objects, an attacker is able to determine which objects are present and to estimate their position in the scene. 
The attack is based on the minimum amount of information that a localization service needs to provide to its users, \ie, camera poses for query images, and exploits that modern localization systems are robust to changing conditions. 
Experiments with our proof-of-concept implementation show the practical feasibility of the attack. 
The attack is applicable even if the localization algorithm used by the server is otherwise perfectly privacy-preserving. 

Our results show that existing privacy-preserving approaches are not sufficient to ensure user privacy, creating the need for further research. 
In particular, first experiments show that preventing the attack proposed in this paper without reducing localization performance and creating other angles of attack is a non-trivial task and interesting direction for future work.

{\small
\noindent\textbf{Acknowledgements. }
This work was supported by the EU Horizon 2020 project RICAIP (grant agreement No. 857306), the European Regional Development Fund under project IMPACT (No. CZ.02.1.01/0.0/0.0/15\_003/0000468), the Czech Science Foundation (GAČR) JUNIOR STAR Grant No. 22-23183M, Chalmers AI Research Center (CHAIR), WASP and SSF.
}%

\bibliographystyle{ieee_fullname}
\bibliography{torsten2}

\begin{thebibliography}{10}\itemsep=-1pt

\bibitem{Brachmann2018CVPR}
Eric Brachmann and Carsten Rother.
\newblock {Learning Less is More - 6D Camera Localization via 3D Surface
  Regression}.
\newblock In {\em CVPR}, 2018.

\bibitem{Brachmann2019ICCVa}
Eric Brachmann and Carsten Rother.
\newblock Expert sample consensus applied to camera re-localization.
\newblock In {\em ICCV}, 2019.

\bibitem{brachmann2020ARXIV}
Eric Brachmann and Carsten Rother.
\newblock Visual camera re-localization from {RGB} and {RGB-D} images using
  {DSAC}.
\newblock {\em arXiv:2002.12324}, 2020.

\bibitem{Camposeco2019CVPR}
Federico Camposeco, Andrea Cohen, Marc Pollefeys, and Torsten Sattler.
\newblock {Hybrid Scene Compression for Visual Localization}.
\newblock In {\em The IEEE Conference on Computer Vision and Pattern
  Recognition (CVPR)}, 2019.

\bibitem{Castle08ISWC}
R.~O. Castle, G. Klein, and D.~W. Murray.
\newblock Video-rate localization in multiple maps for wearable augmented
  reality.
\newblock In {\em ISWC}, 2008.

\bibitem{Cavallari20193DV}
T. {Cavallari}, L. {Bertinetto}, J. {Mukhoti}, P. {Torr}, and S. {Golodetz}.
\newblock Let's take this online: Adapting scene coordinate regression network
  predictions for online rgb-d camera relocalisation.
\newblock In {\em 3DV}, 2019.

\bibitem{Chelani2021CVPR}
Kunal Chelani, Fredrik Kahl, and Torsten Sattler.
\newblock {How Privacy-Preserving Are Line Clouds? Recovering Scene Details
  From 3D Lines}.
\newblock In {\em (CVPR)}, pages 15668--15678, June 2021.

\bibitem{dai2017scannet}
Angela Dai, Angel~X. Chang, Manolis Savva, Maciej Halber, Thomas Funkhouser,
  and Matthias Nie{\ss}ner.
\newblock Scannet: Richly-annotated 3d reconstructions of indoor scenes.
\newblock In {\em CVPR}, 2017.

\bibitem{Dangwal2021ReverseEngineeringPP}
Deeksha Dangwal, Vincent~T. Lee, Hyo~Jin Kim, Tianwei Shen, Meghan Cowan, Rajvi
  Shah, Caroline Trippel, Brandon Reagen, Timothy Sherwood, Vasileios Balntas,
  Armin Alaghi, and Eddy Ilg.
\newblock Analysis and mitigations of reverse engineering attacks on local
  feature descriptors.
\newblock {\em BMVC}, 2021.

\bibitem{detone2017superpoint}
Daniel DeTone, Tomasz Malisiewicz, and Andrew Rabinovich.
\newblock Superpoint: Self-supervised interest point detection and description,
  2018.

\bibitem{Do2022CVPR}
Tien Do, Ondrej Miksik, Joseph DeGol, Hyun~Soo Park, and Sudipta~N. Sinha.
\newblock Learning to detect scene landmarks for camera localization.
\newblock In {\em Proceedings of the IEEE/CVF Conference on Computer Vision and
  Pattern Recognition (CVPR)}, June 2022.

\bibitem{Dosovitskiy2016CVPR}
Alexey Dosovitskiy and Thomas Brox.
\newblock Inverting visual representations with convolutional networks.
\newblock In {\em CVPR 2016}, pages 4829--4837, 06 2016.

\bibitem{Dusmanu2019CVPR}
Mihai Dusmanu, Ignacio Rocco, Tomas Pajdla, Marc Pollefeys, Josef Sivic,
  Akihiko Torii, and Torsten Sattler.
\newblock {D2-Net}: {A} trainable {CNN} for joint detection and description of
  local features.
\newblock In {\em CVPR}, 2019.

\bibitem{Dusmanu2021AdversarialSubspacePP}
Mihai Dusmanu, Johannes~L. Sch{\"{o}}nberger, Sudipta~N. Sinha, and Marc
  Pollefeys.
\newblock Privacy-preserving visual feature descriptors through adversarial
  affine subspace embedding.
\newblock {\em CVPR}, 2021.

\bibitem{Geppert2022CVPR}
Marcel Geppert, Viktor Larsson, Johannes~L. Sch{\"{o}}nberger, and Marc
  Pollefeys.
\newblock Privacy preserving partial localization.
\newblock In {\em CVPR}, 2022.

\bibitem{Geppert2020ECCV}
Marcel Geppert, Viktor Larsson, Pablo Speciale, Johannes~L. Sch{\"{o}}nberger,
  and Marc Pollefeys.
\newblock {Privacy Preserving Structure-from-Motion}.
\newblock In {\em ECCV}, 2020.

\bibitem{Geppert2019ICRA}
Marcel Geppert, Peidong Liu, Zhaopeng Cui, Marc Pollefeys, and Torsten Sattler.
\newblock {Efficient 2D-3D Matching for Multi-Camera Visual Localization}.
\newblock In {\em {ICRA}}, 2019.

\bibitem{Heng2019ICRA}
L. {Heng}, B. {Choi}, Z. {Cui}, M. {Geppert}, S. {Hu}, B. {Kuan}, P. {Liu}, R.
  {Nguyen}, Y.~C. {Yeo}, A. {Geiger}, G.~H. {Lee}, M. {Pollefeys}, and T.
  {Sattler}.
\newblock {Project AutoVision: Localization and 3D Scene Perception for an
  Autonomous Vehicle with a Multi-Camera System}.
\newblock In {\em ICRA}, 2019.

\bibitem{HumenbergerX20Kapture}
Martin Humenberger, Yohann Cabon, Nicolas Guerin, Julien Morat, J\'{e}r\^{o}me
  Revaud, Philippe Rerole, No\'{e} Pion, Cesar de Souza, Vincent Leroy, and
  Gabriela Csurka.
\newblock {Robust Image Retrieval-based Visual Localization using Kapture}.
\newblock arXiv:2007.13867, 2020.

\bibitem{Irschara09CVPR}
A. Irschara, C. Zach, J.-M. Frahm, and H. Bischof.
\newblock {From Structure-from-Motion Point Clouds to Fast Location
  Recognition}.
\newblock In {\em CVPR}, 2009.

\bibitem{Li2012ECCV}
Y. Li, N. Snavely, D. Huttenlocher, and P. Fua.
\newblock {Worldwide Pose Estimation Using 3D Point Clouds}.
\newblock In {\em ECCV}, 2012.

\bibitem{Lowe04IJCV}
D. Lowe.
\newblock {Distinctive Image Features from Scale-Invariant Keypoints}.
\newblock {\em {IJCV}}, 60(2), 2004.

\bibitem{Lynen2020IJRR}
Simon Lynen, Bernhard Zeisl, Dror Aiger, Michael Bosse, Joel Hesch, Marc
  Pollefeys, Roland Siegwart, and Torsten Sattler.
\newblock Large-scale, real-time visual–inertial localization revisited.
\newblock {\em {IJRR}}, 39(9):1061--1084, 2020.

\bibitem{MicrosoftSpatialAnchors}
Microsoft.
\newblock {Spatial Anchors}, 2020.

\bibitem{Middelberg2014ECCV}
S. Middelberg, T. Sattler, O. Untzelmann, and L. Kobbelt.
\newblock {Scalable 6-DOF Localization on Mobile Devices}.
\newblock In {\em {ECCV}}, 2014.

\bibitem{Ng2021NinjaDescPP}
Tony Ng, Hyo~Jin Kim, Vincent~T. Lee, Daniel DeTone, Tsun{-}Yi Yang, Tianwei
  Shen, Eddy Ilg, Vassileios Balntas, Krystian Mikolajczyk, and Chris Sweeney.
\newblock Ninjadesc: Content-concealing visual descriptors via adversarial
  learning.
\newblock {\em CVPR}, 2022.

\bibitem{Panek2022ECCV}
Vojtech Panek, Zuzana Kukelova, and Torsten Sattler.
\newblock {MeshLoc: Mesh-Based Visual Localization}.
\newblock In {\em ECCV}, 2022.

\bibitem{pittaluga2019revealing}
Francesco Pittaluga, Sanjeev~J Koppal, Sing Bing~Kang, and Sudipta~N Sinha.
\newblock Revealing scenes by inverting structure from motion reconstructions.
\newblock In {\em CVPR}, pages 145--154, 2019.

\bibitem{GoogleVPS}
Tilman Reinhardt.
\newblock {Using Global Localization to Improve Navigation}, 2019.

\bibitem{r2d2}
Jerome Revaud, Philippe Weinzaepfel, C{\'{e}}sar~Roberto de Souza, and Martin
  Humenberger.
\newblock {R2D2:} repeatable and reliable detector and descriptor.
\newblock In {\em NeurIPS}, 2019.

\bibitem{sarlin2019coarse}
Paul-Edouard Sarlin, Cesar Cadena, Roland Siegwart, and Marcin Dymczyk.
\newblock From coarse to fine: Robust hierarchical localization at large scale.
\newblock In {\em CVPR}, 2019.

\bibitem{sarlin20superglue}
Paul-Edouard Sarlin, Daniel DeTone, Tomasz Malisiewicz, and Andrew Rabinovich.
\newblock {SuperGlue}: Learning feature matching with graph neural networks.
\newblock In {\em CVPR}, 2020.

\bibitem{Sattler2017PAMI}
T. Sattler, B. Leibe, and L. Kobbelt.
\newblock {Efficient \& Effective Prioritized Matching for Large-Scale
  Image-Based Localization}.
\newblock {\em PAMI}, 39(9):1744--1756, 2017.

\bibitem{Schoenberger2016CVPR}
Johannes~L. Sch\"{o}nberger and Jan-Michael Frahm.
\newblock {Structure-From-Motion Revisited}.
\newblock In {\em CVPR}, June 2016.

\bibitem{Schoenberger2016ECCV}
Johannes~Lutz Sch\"{o}nberger, Enliang Zheng, Marc Pollefeys, and Jan-Michael
  Frahm.
\newblock {Pixelwise View Selection for Unstructured Multi-View Stereo}.
\newblock In {\em European Conference on Computer Vision (ECCV)}, 2016.

\bibitem{Shibuya2020ECCV}
Mikiya Shibuya, Shinya Sumikura, and Ken Sakurada.
\newblock {Privacy Preserving Visual {SLAM}}.
\newblock In {\em ECCV}, 2020.

\bibitem{shibuya2020privacy}
Mikiya Shibuya, Shinya Sumikura, and Ken Sakurada.
\newblock Privacy preserving visual {SLAM}.
\newblock In {\em (ECCV)}, 2020.

\bibitem{Shotton2013CVPR}
Jamie Shotton, Ben Glocker, Christopher Zach, Shahram Izadi, Antonio Criminisi,
  and Andrew Fitzgibbon.
\newblock {Scene Coordinate Regression Forests for Camera Relocalization in
  RGB-D Images}.
\newblock In {\em CVPR}, 2013.

\bibitem{niantic_lightship}
Tory Smith.
\newblock Niantic lightship.
\newblock 2022.

\bibitem{Song2020ECCV}
Zhenbo Song, Wayne Chen, Dylan Campbell, and Hongdong Li.
\newblock {Deep Novel View Synthesis from Colored 3D Point Clouds}.
\newblock In {\em ECCV}, 2020.

\bibitem{speciale2019CVPR}
Pablo Speciale, Sing~Bing Kang, Marc Pollefeys, Johannes Schönberger, and
  Sudipta Sinha.
\newblock Privacy preserving image-based localization.
\newblock In {\em CVPR}. IEEE, June 2019.

\bibitem{Speciale2019ICCV}
Pablo Speciale, Johannes~L. Schonberger, Sudipta~N. Sinha, and Marc Pollefeys.
\newblock {Privacy Preserving Image Queries for Camera Localization}.
\newblock In {\em The IEEE International Conference on Computer Vision (ICCV)},
  2019.

\bibitem{sun2021loftr}
Jiaming Sun, Zehong Shen, Yuang Wang, Hujun Bao, and Xiaowei Zhou.
\newblock {LoFTR}: Detector-free local feature matching with transformers.
\newblock {\em CVPR}, 2021.

\bibitem{Toft2021PAMI}
Carl Toft, Will Maddern, Akihiko Torii, Lars Hammarstrand, Erik Stenborg,
  Daniel Safari, Masatoshi Okutomi, Marc Pollefeys, Josef Sivic, Tomas Pajdla,
  Fredrik Kahl, and Torsten Sattler.
\newblock Long-term visual localization revisited.
\newblock {\em PAMI}, 2020.

\bibitem{Valentin20163DV}
J. {Valentin}, A. {Dai}, M. {Niessner}, P. {Kohli}, P. {Torr}, S. {Izadi}, and
  C. {Keskin}.
\newblock {Learning to Navigate the Energy Landscape}.
\newblock In {\em International Conference on 3D Vision (3DV)}, 2016.

\bibitem{Ventura2014TVCG}
Jonathan Ventura, Clemens Arth, Gerhard Reitmayr, and Dieter Schmalstieg.
\newblock {Global Localization from Monocular SLAM on a Mobile Phone}.
\newblock {\em IEEE Transactions on Visualization and Computer Graphics},
  20(4):531--539, 2014.

\bibitem{wald2020}
Johanna Wald, Torsten Sattler, Stuart Golodetz, Tommaso Cavallari, and Federico
  Tombari.
\newblock Beyond controlled environments: 3d camera re-localization in changing
  indoor scenes.
\newblock In {\em European Conference on Computer Vision (ECCV)}, 2020.

\bibitem{wang2022Matchformer}
Qing Wang, Jiaming Zhang, Kailun Yang, Kunyu Peng, and Rainer Stiefelhagen.
\newblock Matchformer: Interleaving attention in transformers for feature
  matching, 2022.

\bibitem{Weinzaepfel2011CVPR}
P. {Weinzaepfel}, H. {Jégou}, and P. {Pérez}.
\newblock Reconstructing an image from its local descriptors.
\newblock In {\em CVPR}, 2011.

\bibitem{Wendel11ICRA}
A. Wendel, A. Irschara, and H. Bischof.
\newblock Natural landmark-based monocular localization for mavs.
\newblock In {\em ICRA}, 2011.

\bibitem{Yang2022CVPR}
Luwei Yang, Rakesh Shrestha, Wenbo Li, Shuaicheng Liu, Guofeng Zhang, Zhaopeng
  Cui, and Ping Tan.
\newblock Scenesqueezer: Learning to compress scene for camera relocalization.
\newblock In {\em Proceedings of the IEEE/CVF Conference on Computer Vision and
  Pattern Recognition (CVPR)}, pages 8259--8268, June 2022.

\bibitem{Zhou2022ECCV}
Qunjie Zhou, S{\'e}rgio Agostinho, Aljo{\v{s}}a O{\v{s}}ep, and Laura
  Leal-Taix{\'e}.
\newblock Is geometry enough for matching in visual localization?
\newblock In {\em ECCV}, 2022.

\bibitem{Zhou2021CVPR}
Qunjie Zhou, Torsten Sattler, and Laura Leal-Taixe.
\newblock Patch2pix: Epipolar-guided pixel-level correspondences.
\newblock In {\em Proceedings of the IEEE/CVF Conference on Computer Vision and
  Pattern Recognition (CVPR)}, 2021.

\end{thebibliography}

\appendix
\appendixpage

In Section \ref{sec:qual_ikea}, we present additional qualitative results for  alignments of objects from the \textit{IKEA-Objects} set in all scenes from the \textit{IKEA-Scenes} dataset. Section \ref{sec:qual_rio} presents qualitative results for object alignments in some of the RIO10 scenes, corresponding to the quantitative results shown in Figure 6 of the main paper.


\section{Qualitative results - \textit{Ikea-Scenes} and \textit{Ikea-Objects}}\label{sec:qual_ikea} 

\begin{figure*}[t!]
    \centering
    \includegraphics[width=\linewidth]{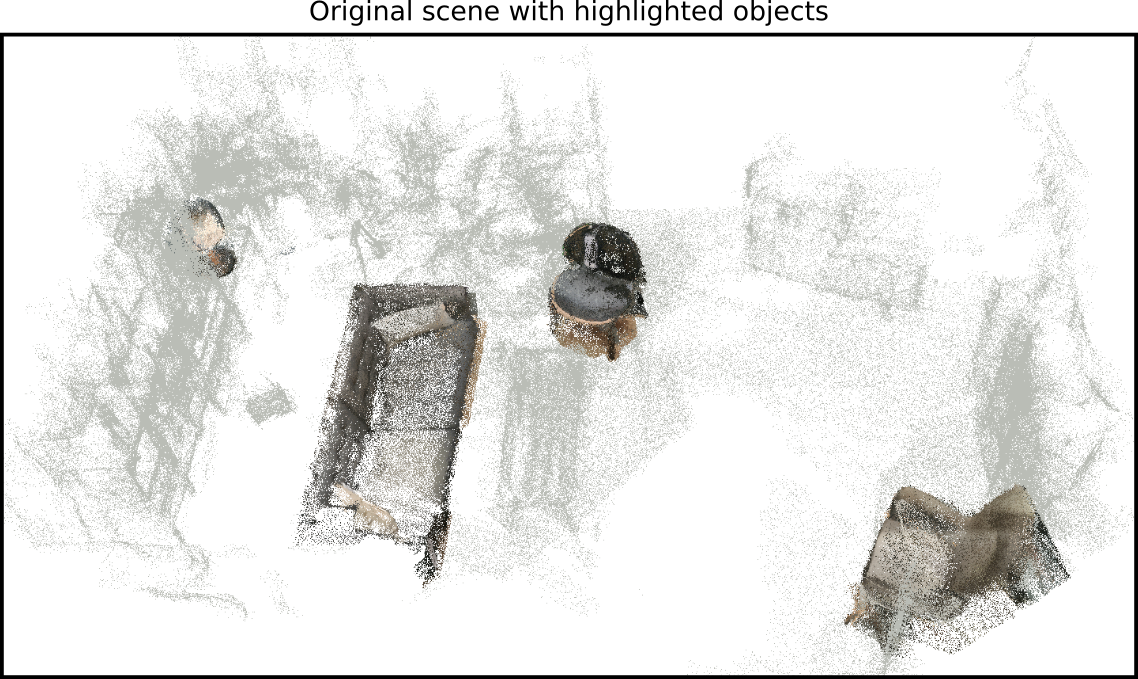}
    \caption{Scene01 of \textit{IKEA-Scenes} with selected objects in focus.}
    \label{fig:ikea_scene01_overview}
\end{figure*}

\begin{figure*}[t!]
    \centering
    \includegraphics[width=\linewidth]{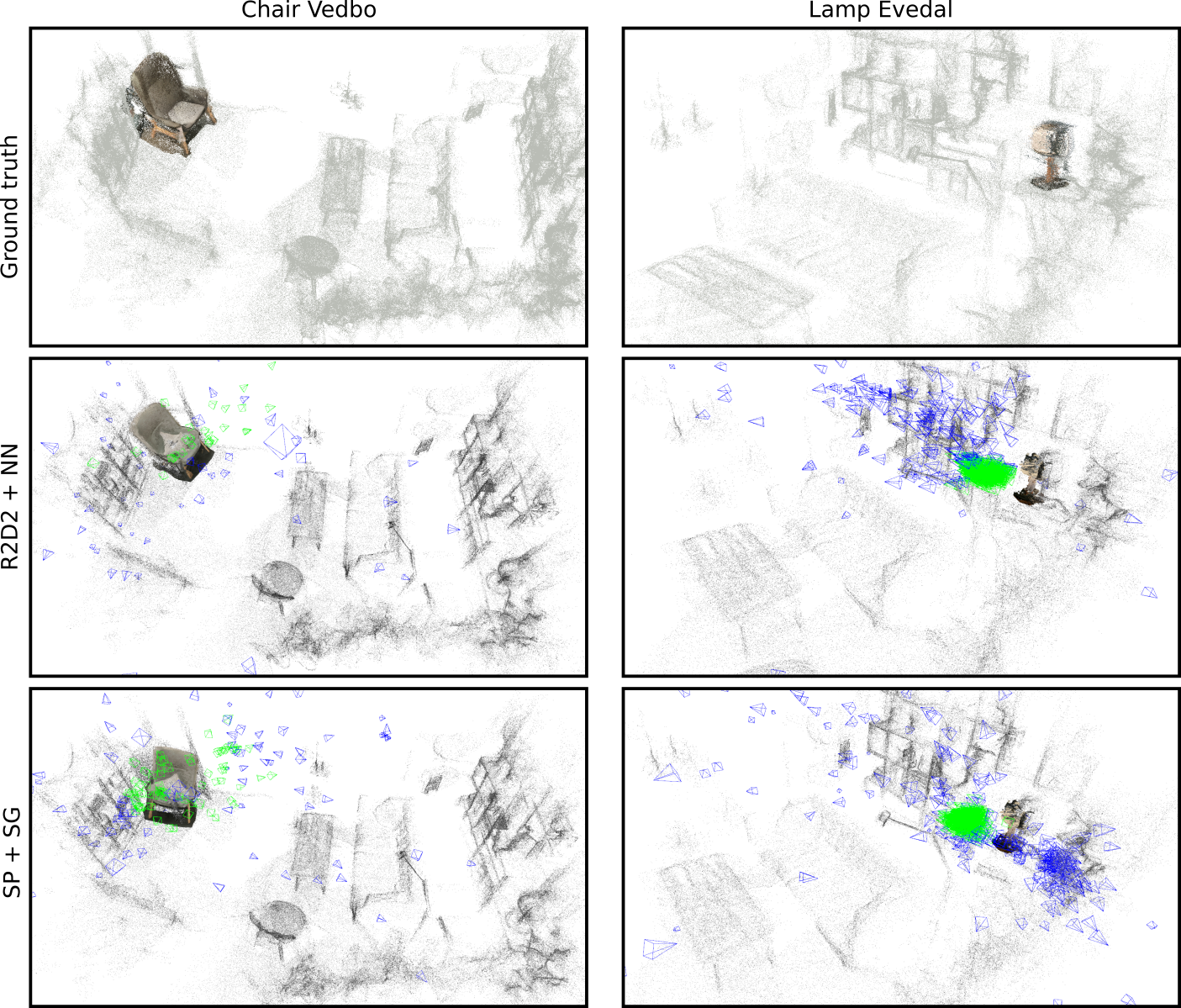}
    \caption{Qualitative results for aligning corresponding objects from \textit{IKEA-Objects} using poses for localizing them in  Scene01 of \textit{IKEA-Scenes}.}
    \label{fig:ikea_scene01_align1}
\end{figure*}

\begin{figure*}[t!]
    \centering
    \includegraphics[width=\linewidth]{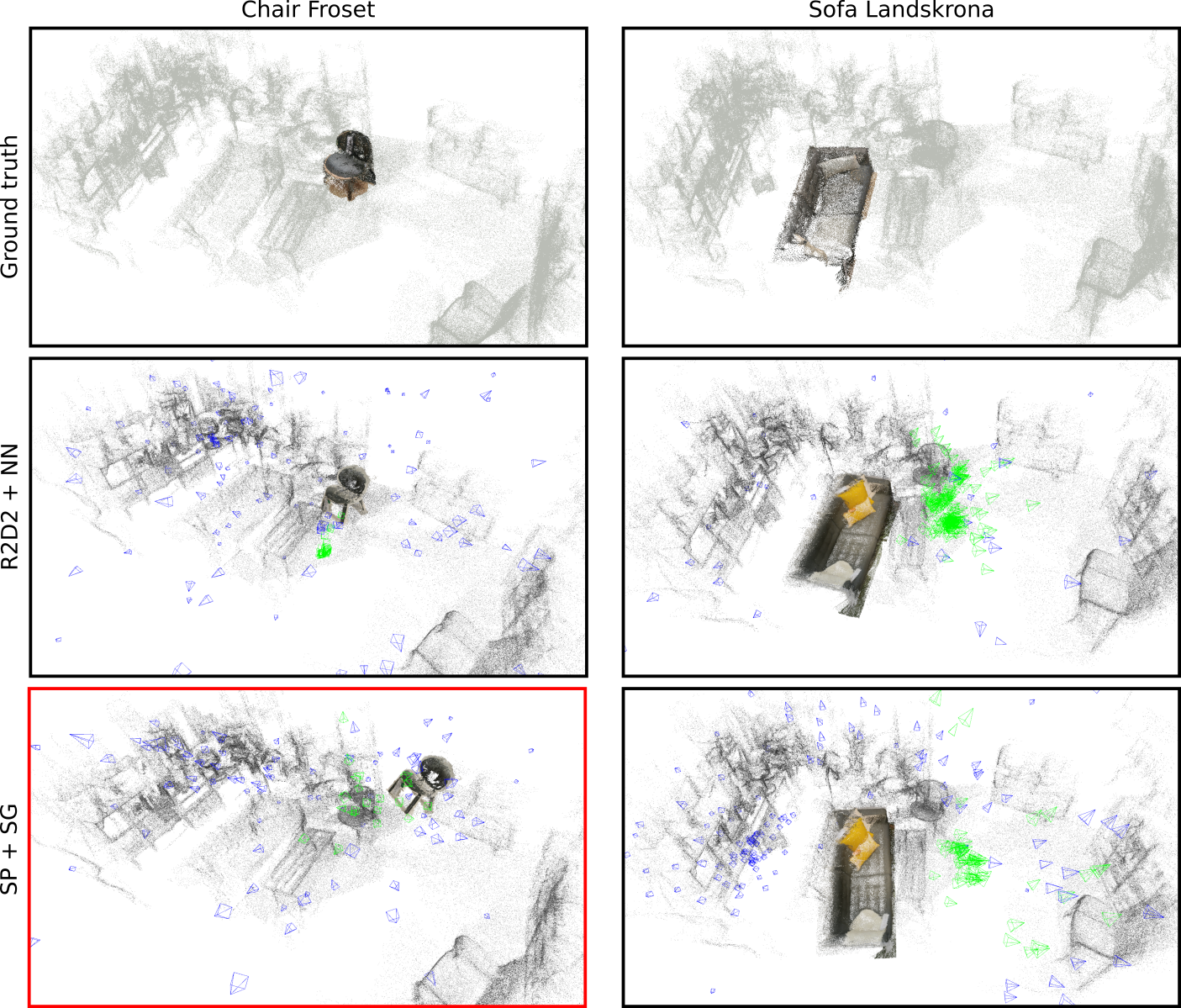}
    \caption{Qualitative results for aligning corresponding objects from \textit{IKEA-Objects} using poses for localizing them in  Scene01 of \textit{IKEA-Scenes}. \textbf{Failure case}: The aligned model for Chair Froset when using Superpoint+Superglue is a bit far from the actual object and also oriented incorrectly.}
    \label{fig:ikea_scene01_align2}
\end{figure*}

\begin{figure*}[t!]
    \centering
    \includegraphics[width=\linewidth]{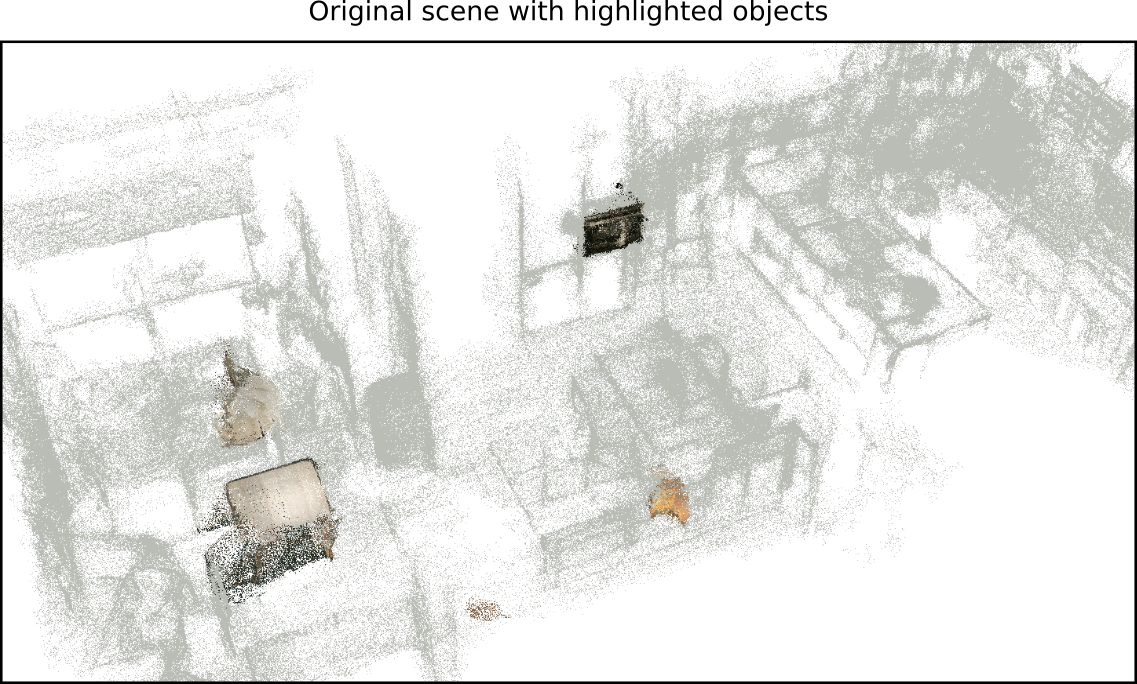}
    \caption{Scene02 of \textit{IKEA-Scenes} with selected objects in focus.}
    \label{fig:ikea_scene02_overview}
\end{figure*}

\begin{figure*}[t!]
    \centering
    \includegraphics[width=\linewidth]{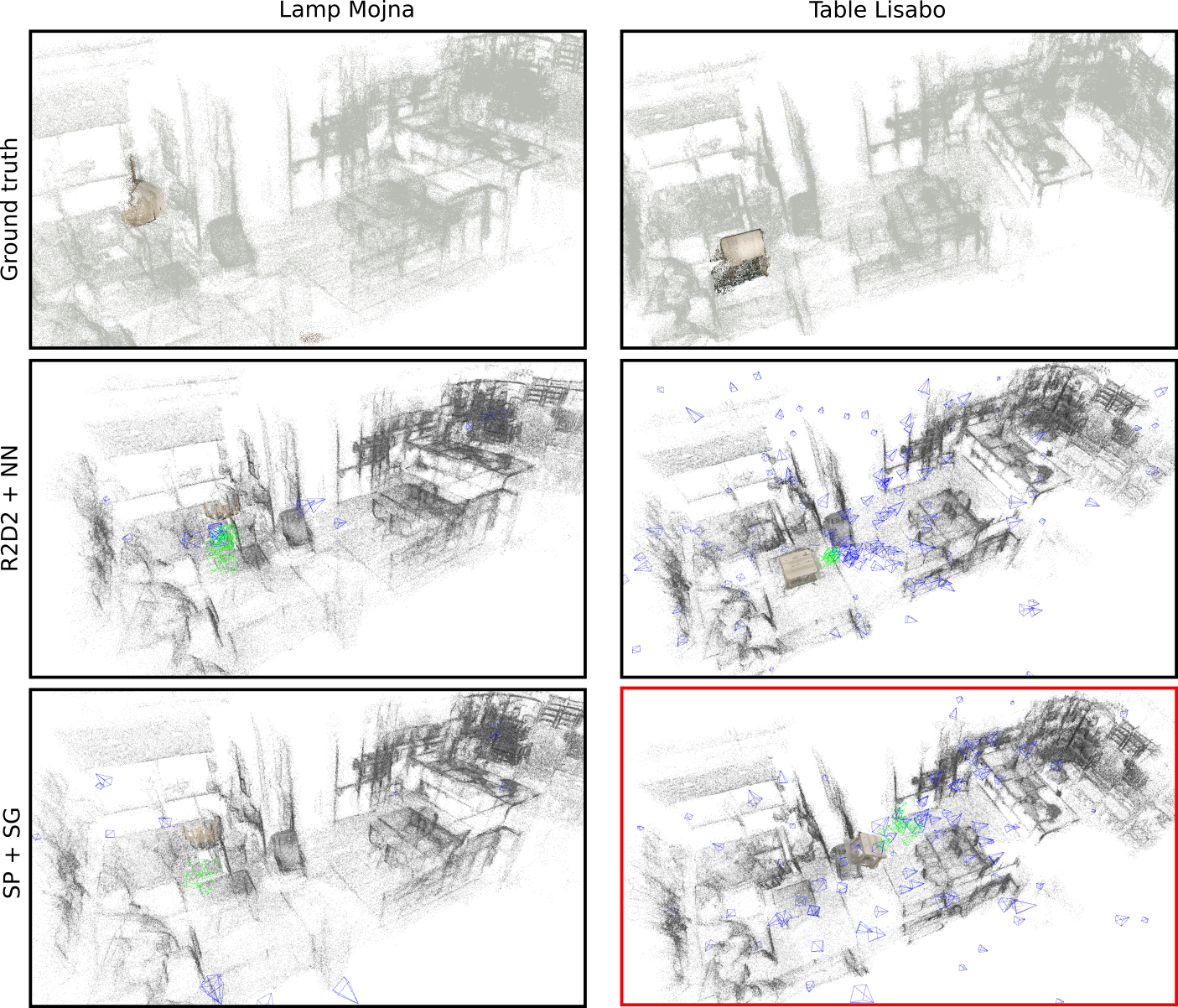}
    \caption{Qualitative results for aligning corresponding objects from \textit{IKEA-Objects} using poses for localizing them in  Scene02 of \textit{IKEA-Scenes}. \textbf{Failure case}: The aligned model for Table Lisabo when using Superpoint+Superglue is a bit far from the actual object and also oriented incorrectly.}
    \label{fig:ikea_scene02_align1}
\end{figure*}

\begin{figure*}[t!]
    \centering
    \includegraphics[width=\linewidth]{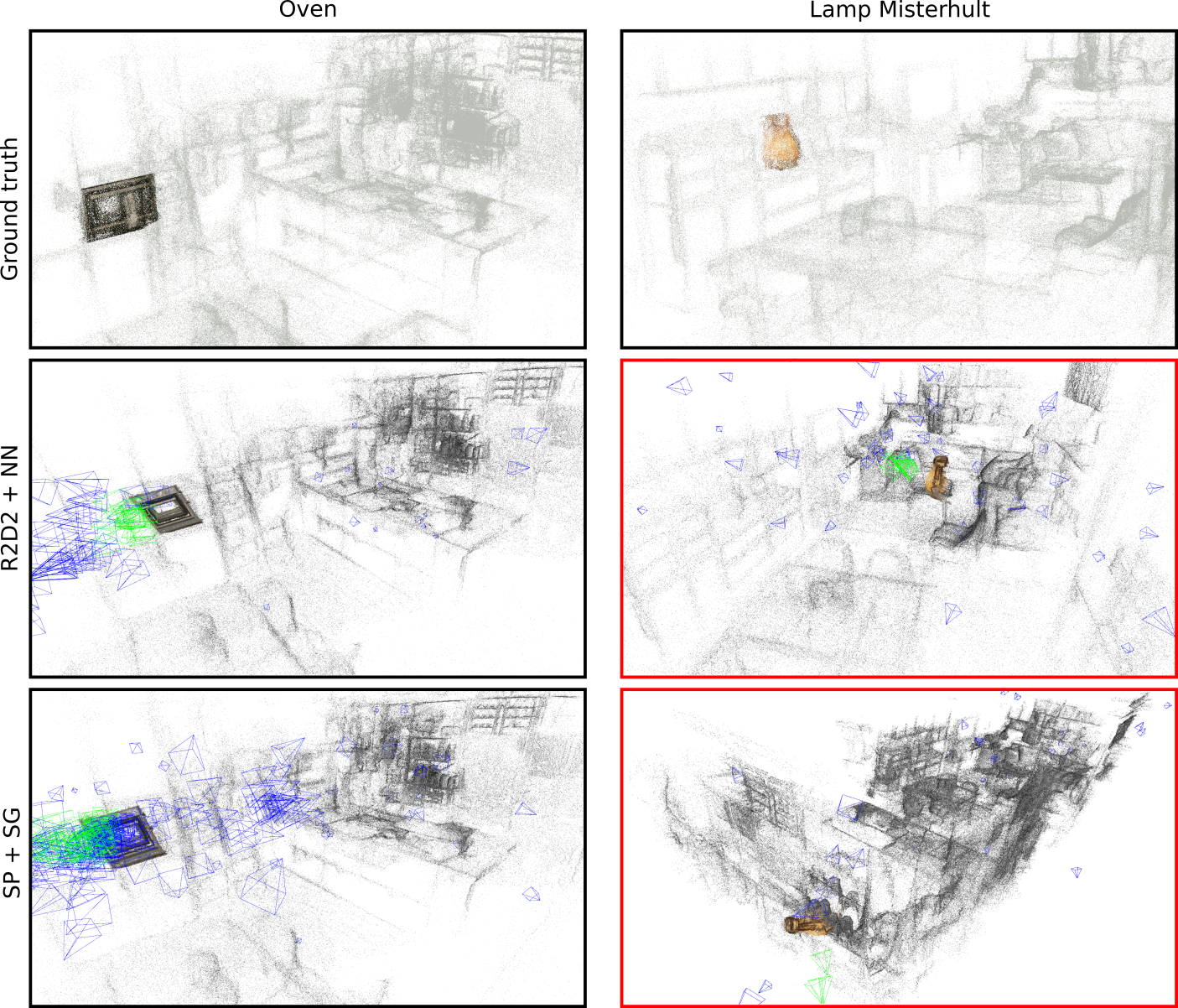}
    \caption{Qualitative results for aligning corresponding objects from \textit{IKEA-Objects} using poses for localizing them in  Scene02 of \textit{IKEA-Scenes}. \textbf{Failure cases}: The aligned models of Lamp Misterhult when using R2D2+NN or Superpoint+Superglue are very far from the actual object. This can be attributed to the similar wooden  appearance of the lamp and several wooden objects in the scene, which leads to incorrect matches.}
    \label{fig:ikea_scene02_align2}
\end{figure*}

\begin{figure*}[t!]
    \centering
    \includegraphics[width=\linewidth]{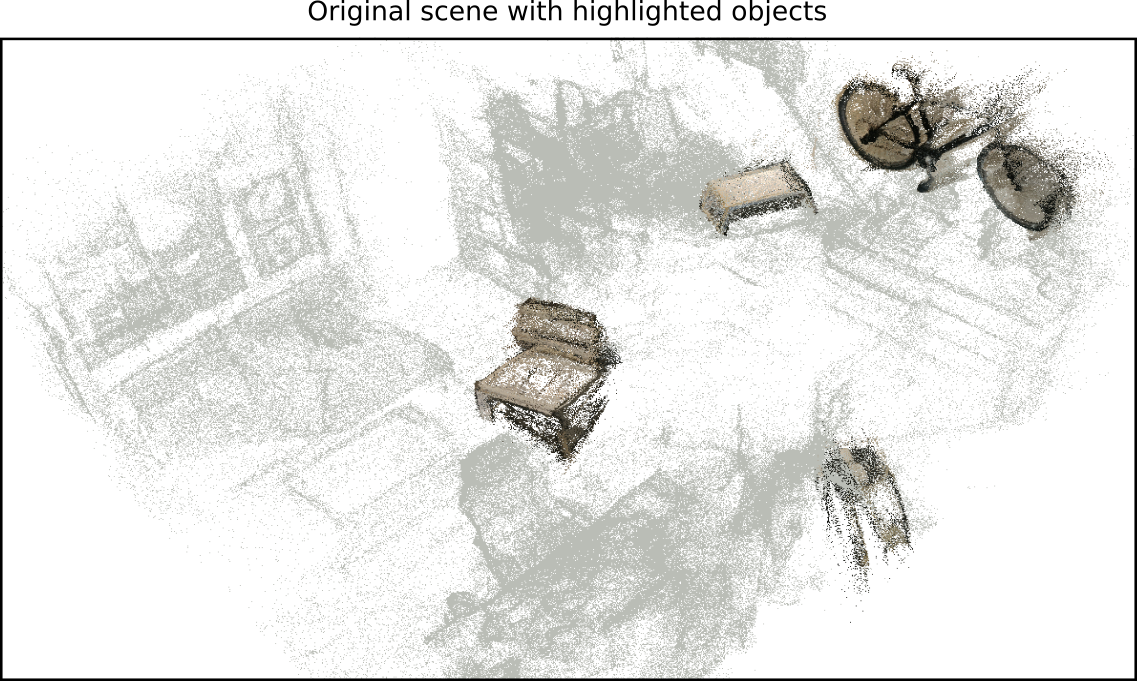}
    \caption{Scene03 of \textit{IKEA-Scenes} with selected objects in focus.}
    \label{fig:ikea_scene03_overview}
\end{figure*}

\begin{figure*}[t!]
    \centering
    \includegraphics[width=\linewidth]{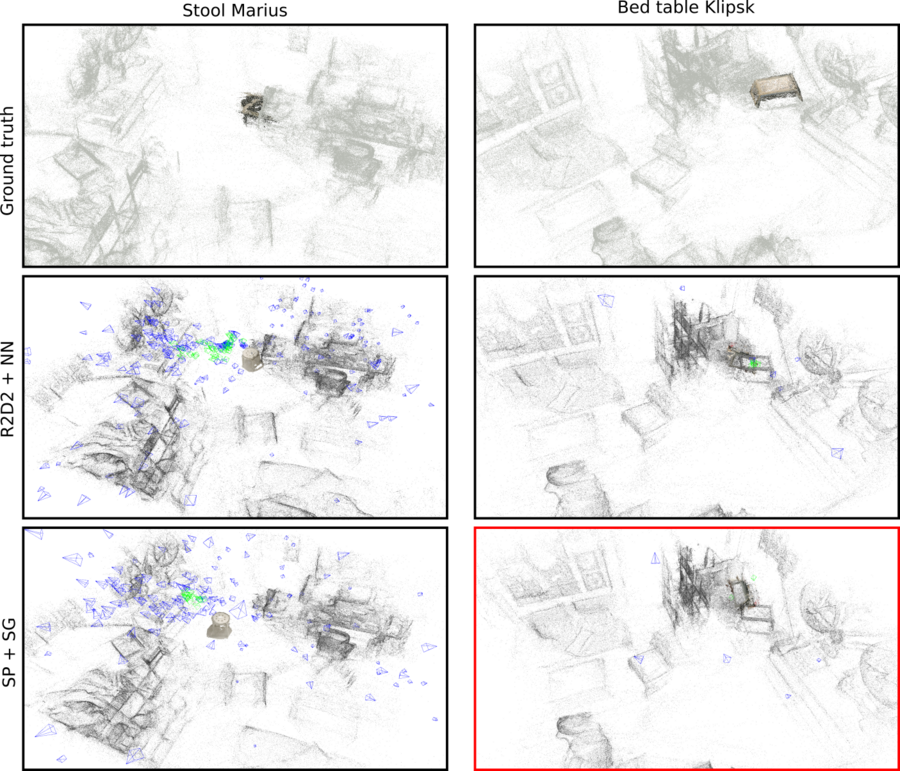}
    \caption{Qualitative results for aligning corresponding objects from \textit{IKEA-Objects} using poses for localizing them in  Scene03 of \textit{IKEA-Scenes}. \textbf{Failure case}: The aligned model for Bed table Klipsk when using Superpoint+Superglue is close to the actual object, but incorrectly oriented.}
    \label{fig:ikea_scene03_align1}
\end{figure*}

\begin{figure*}[t!]
    \centering
    \includegraphics[width=\linewidth]{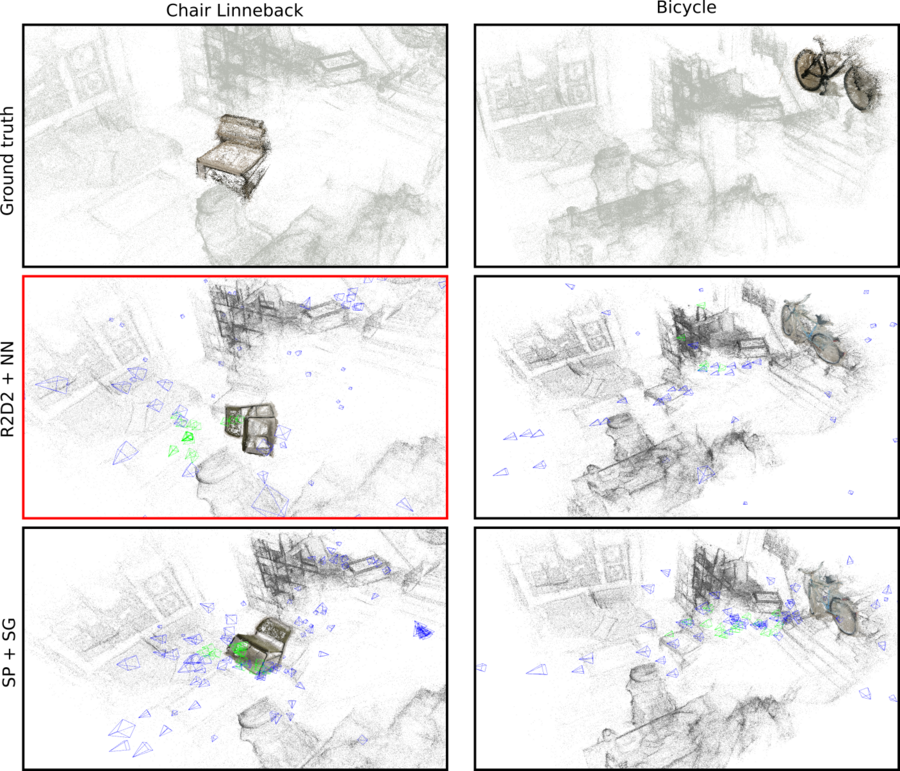}
    \caption{Qualitative results for aligning corresponding objects from \textit{IKEA-Objects} using poses for localizing them in  Scene03 of \textit{IKEA-Scenes}. \textbf{Failure case}: The aligned model for Chair Linneback when using R2D2+NN is close to the actual object, but incorrectly oriented.}
    \label{fig:ikea_scene03_align2}
\end{figure*}

\begin{figure*}[t!]
    \centering
    \includegraphics[width=\linewidth]{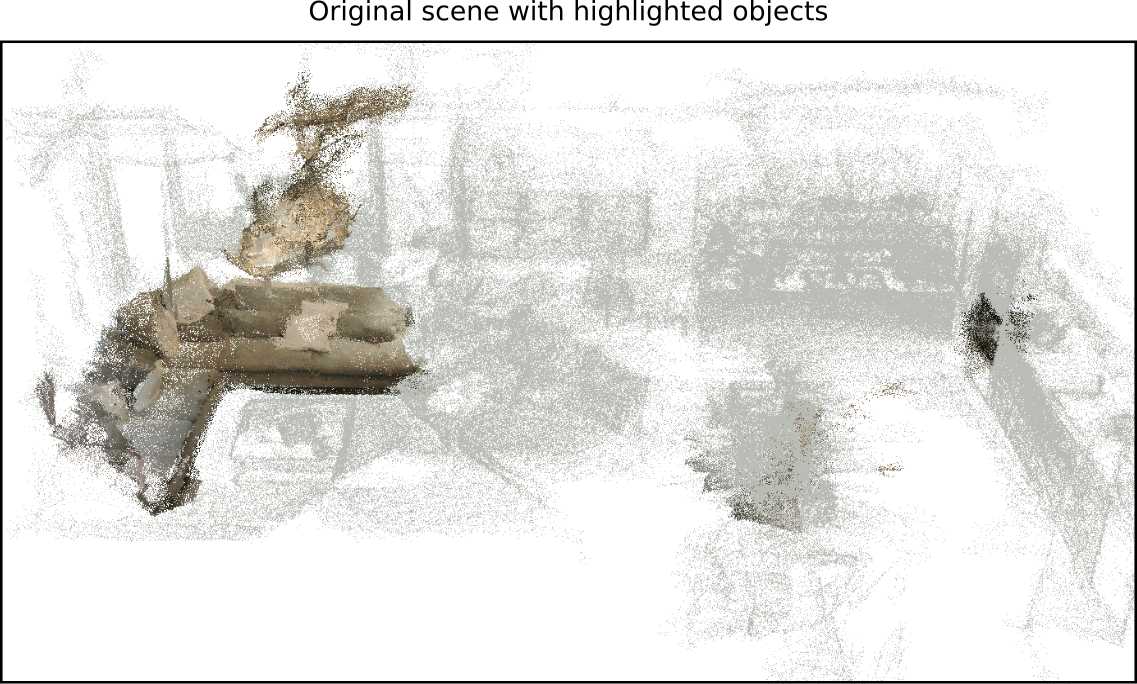}
    \caption{Scene04 of \textit{IKEA-Scenes} with selected objects in focus.}
    \label{fig:ikea_scene04_overview}
\end{figure*}

\begin{figure*}[t!]
    \centering
    \includegraphics[width=\linewidth]{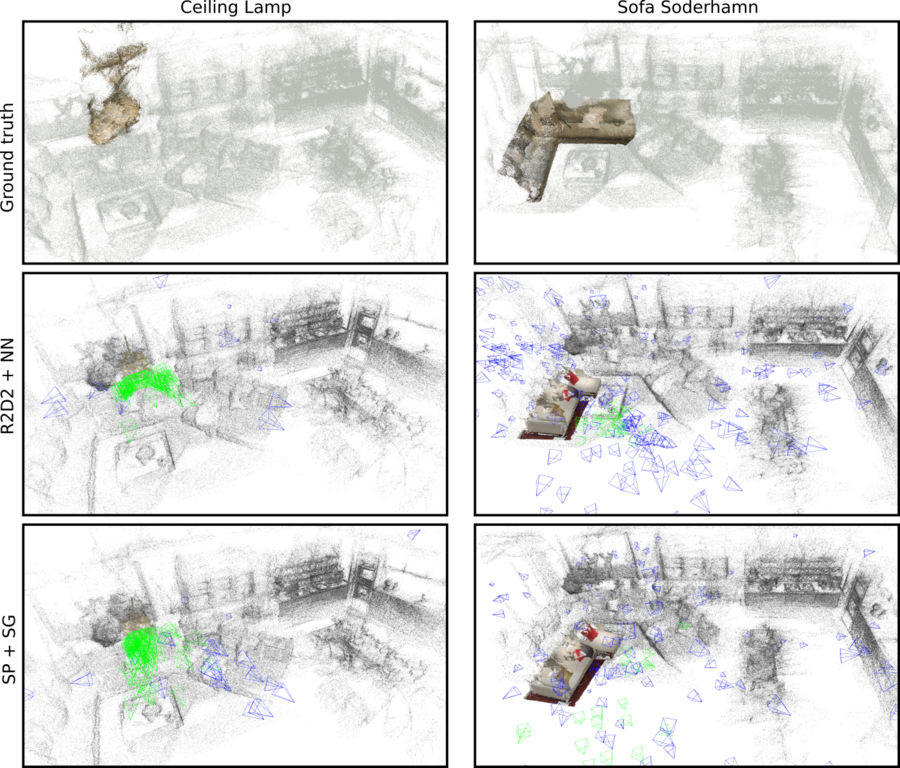}
    \caption{Qualitative results for aligning corresponding objects from \textit{IKEA-Objects} using poses for localizing them in  Scene04 of \textit{IKEA-Scenes}. }
    \label{fig:ikea_scene04_align1}
\end{figure*}

\begin{figure*}[t!]
    \centering
    \includegraphics[width=\linewidth]{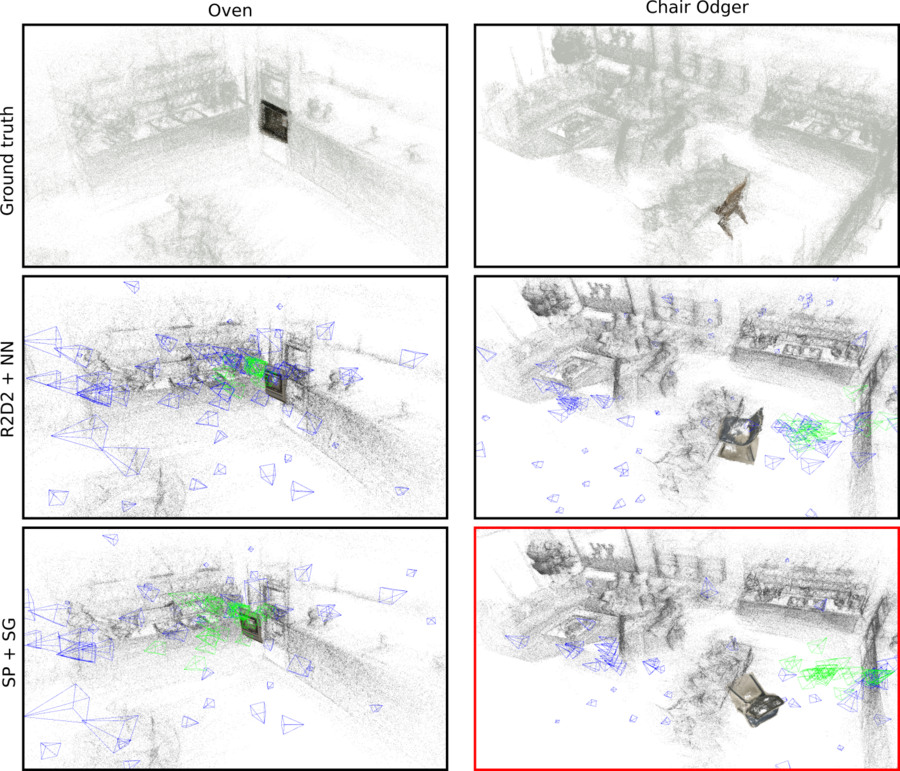}
    \caption{Qualitative results for aligning corresponding objects from \textit{IKEA-Objects} using poses for localizing them in  Scene04 of \textit{IKEA-Scenes}. \textbf{Failure case}: The aligned model for Chair Odger when using Superpoint+Superglue is close to the actual object, but incorrectly oriented.}
    \label{fig:ikea_scene04_align2}
\end{figure*}

\begin{figure*}[t!]
    \centering
    \includegraphics[width=\linewidth]{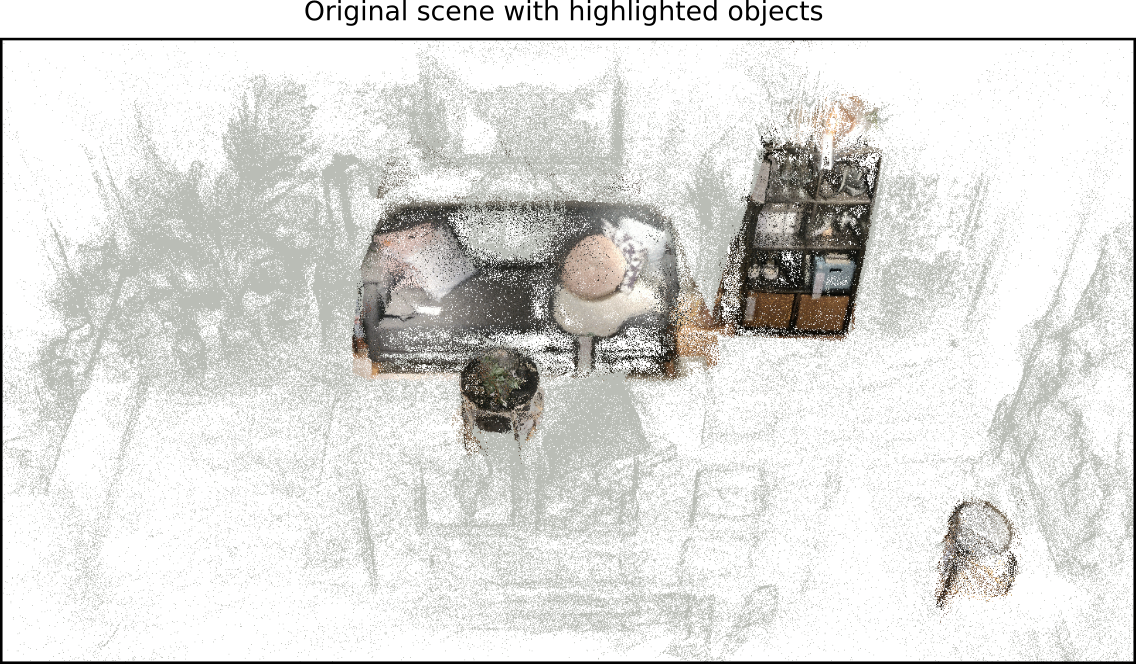}
    \caption{Scene05 of \textit{IKEA-Scenes} with selected objects in focus.}
    \label{fig:ikea_scene05_overview}
\end{figure*}

\begin{figure*}[t!]
    \centering
    \includegraphics[width=\linewidth]{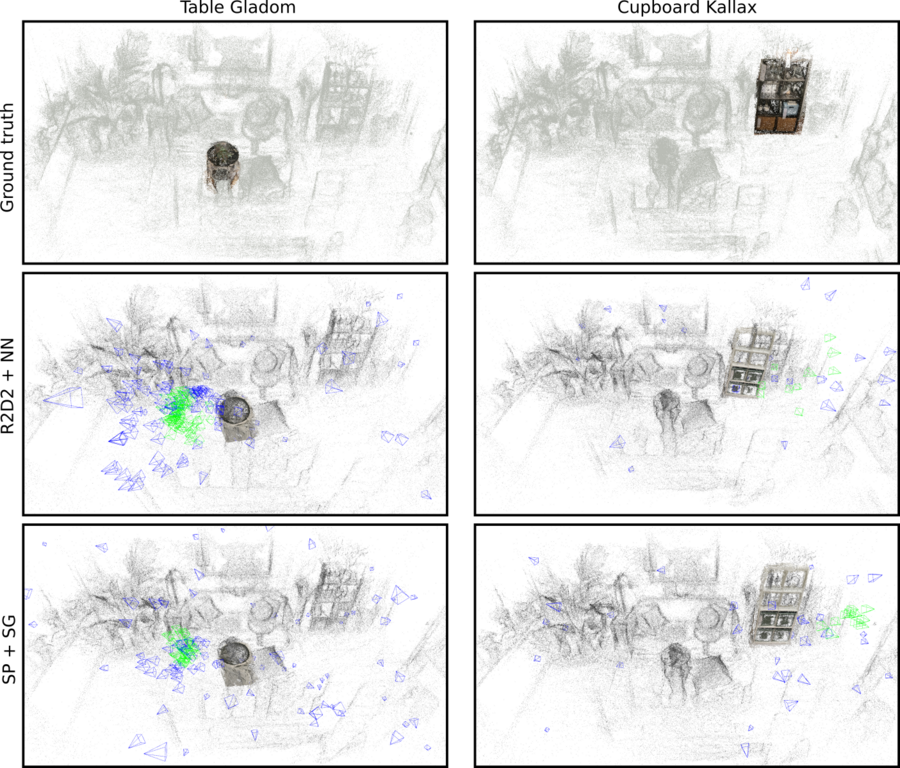}
    \caption{Qualitative results for aligning corresponding objects from \textit{IKEA-Objects} using poses for localizing them in Scene05 of \textit{IKEA-Scenes}.}
    \label{fig:ikea_scene05_align1}
\end{figure*}

\begin{figure*}[t!]
    \centering
    \includegraphics[width=\linewidth]{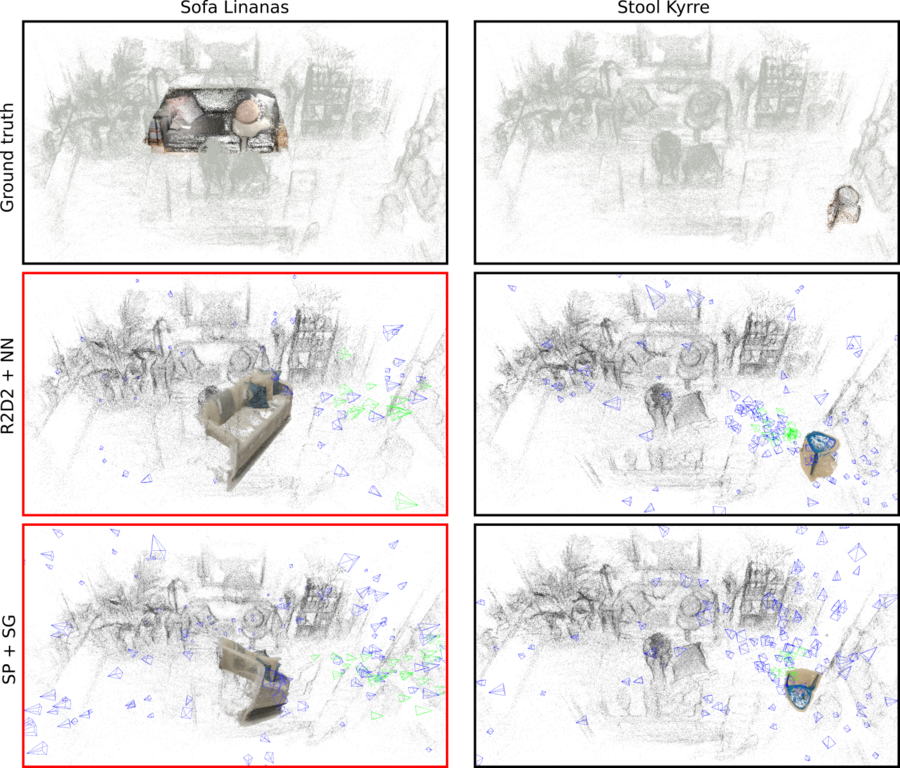}
    \caption{Qualitative results for aligning corresponding objects from \textit{IKEA-Objects} using poses for localizing them in Scene05 of \textit{IKEA-Scenes}. \textbf{Failure cases}: The aligned model for Sofa Linanas when using Superpoint+Superglue or R2D2+NN is incorrectly oriented and also far from the actual object.}
    \label{fig:ikea_scene05_align2}
\end{figure*}

\begin{figure*}[t!]
    \centering
    \includegraphics[width=\linewidth]{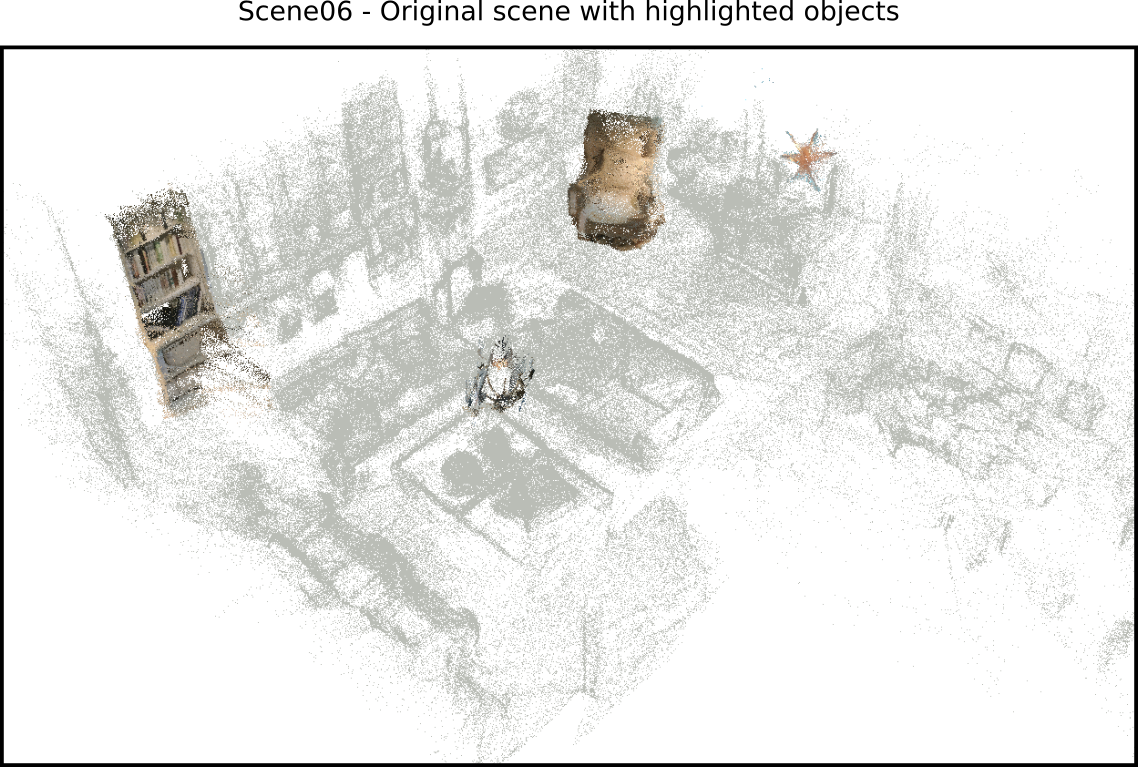}
    \caption{Scene06 of \textit{IKEA-Scenes} with selected objects in focus.}
    \label{fig:ikea_scene06_overview}
\end{figure*}

\begin{figure*}[t!]
    \centering
    \includegraphics[width=\linewidth]{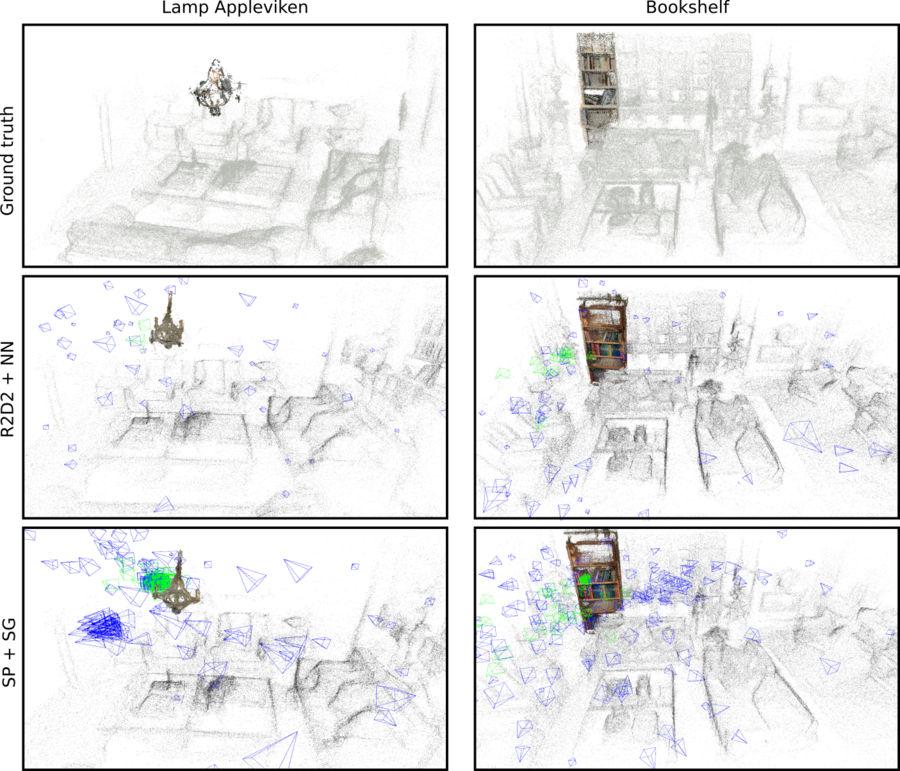}
    \caption{Qualitative results for aligning corresponding objects from \textit{IKEA-Objects} using poses for localizing them in Scene06 of \textit{IKEA-Scenes}.}
    \label{fig:ikea_scene06_align1}
\end{figure*}

\begin{figure*}[t!]
    \centering
    \includegraphics[width=\linewidth]{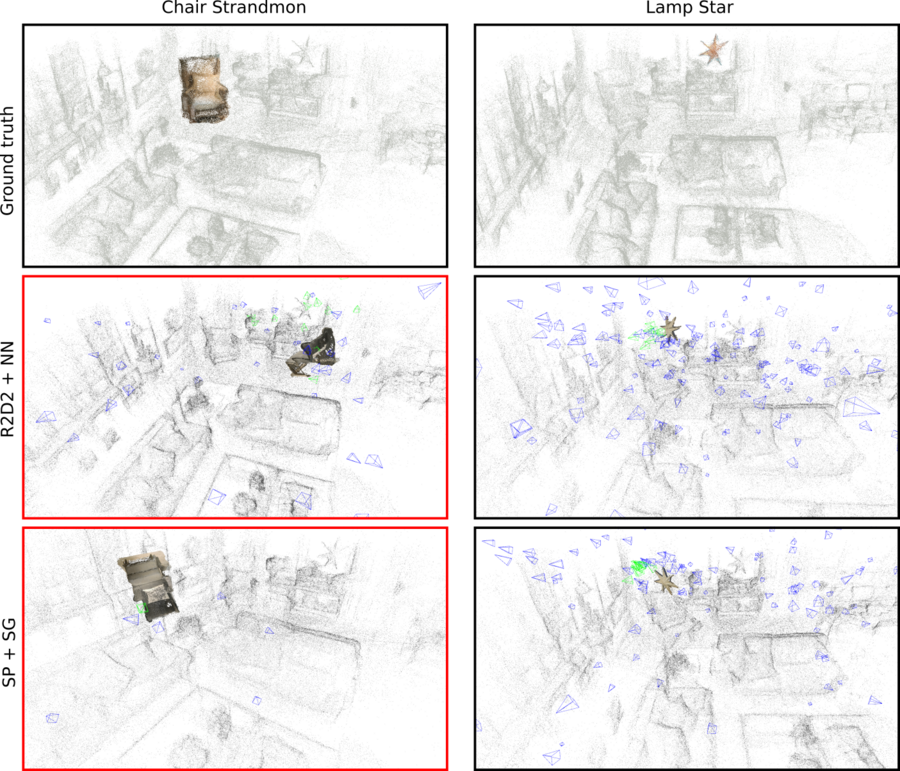}
    \caption{Qualitative results for aligning corresponding objects from \textit{IKEA-Objects} using poses for localizing them in Scene06 of \textit{IKEA-Scenes}. \textbf{Failure cases}: The aligned model for Chair Strandmon when using Superpoint+Superglue or R2D2+NN is incorrectly aligned and also far from the actual object.}
    \label{fig:ikea_scene06_align2}
\end{figure*}

\begin{figure*}[t!]
    \centering
    \includegraphics[width=\linewidth]{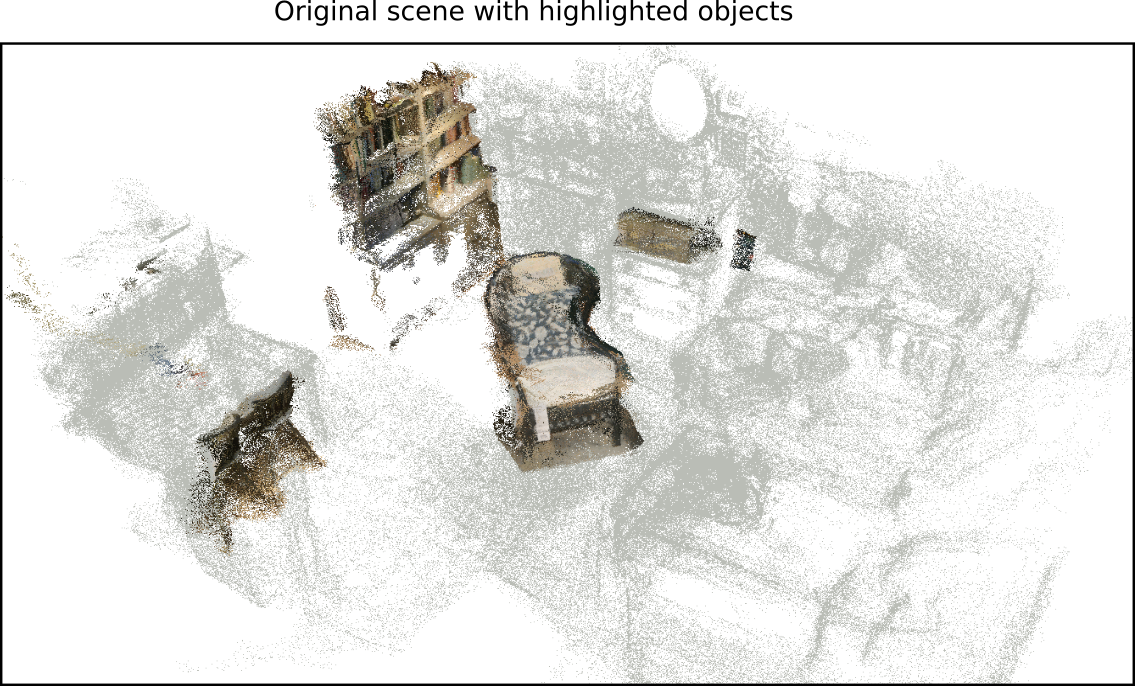}
    \caption{Scene07 of \textit{IKEA-Scenes} with selected objects in focus.}
    \label{fig:ikea_scene07_overview}
\end{figure*}

\begin{figure*}[t!]
    \centering
    \includegraphics[width=\linewidth]{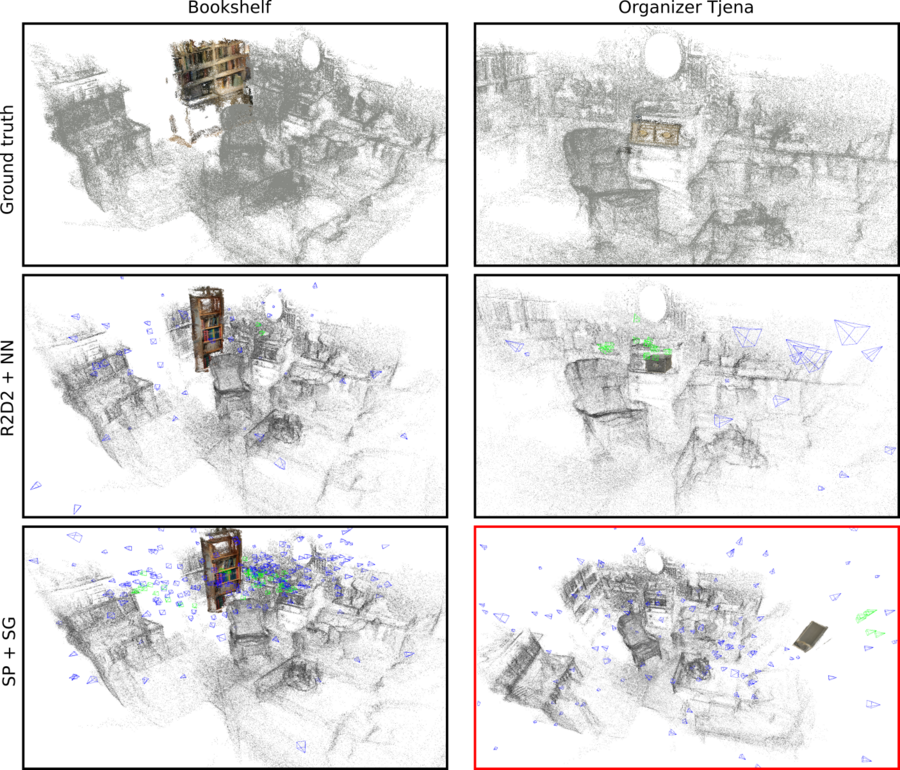}
    \caption{Qualitative results for aligning corresponding objects from \textit{IKEA-Objects} using poses for localizing them in Scene07 of \textit{IKEA-Scenes}. \textbf{Failure case}: The aligned model for Organizer Tjena when using Superpoint+Superglue is far from the actual object and also incorrectly oriented.}
    \label{fig:ikea_scene07_align1}
\end{figure*}

\begin{figure*}[t!]
    \centering
    \includegraphics[width=\linewidth]{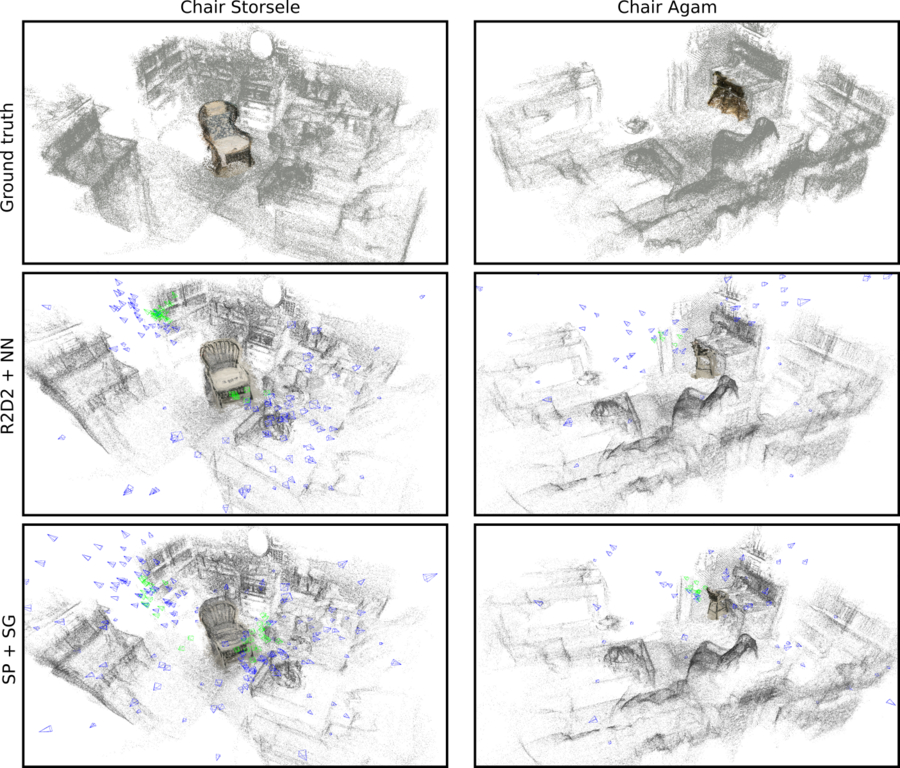}
    \caption{Qualitative results for aligning corresponding objects from \textit{IKEA-Objects} using poses for localizing them in Scene07 of \textit{IKEA-Scenes}.}
    \label{fig:ikea_scene07_align2}
\end{figure*}

\PAR{Qualitative alignment results.} 
In Figures \ref{fig:ikea_scene01_overview}-\ref{fig:ikea_scene07_align2}, we present alignment results for four selected objects present in each scene from the the \textit{IKEA-Scenes} dataset. We include results for poses obtained by using 1) Superpoint~\cite{detone2017superpoint} features with Superglue~\cite{sarlin20superglue}-based matching and 2) R2D2~\cite{r2d2} features with Nearest Neighbor matching within the Hloc~\cite{sarlin2019coarse,sarlin20superglue} pipeline.
We selected objects of varying sizes, shapes, categories, textures \etc 
as to show the feasibility of the attack. 
As can be seen, it is often possible to quite accurately place objects in a scene based on camera poses estimated by a visual localization system.

\PAR{Camera poses.} 
The same figures also show the set of camera poses returned by the server (blue) and the subset of poses (green) selected as inliers by our alignment method (see Algorithm 1 and Section 3.1 in the main paper). 
Note that these poses are obtained by localizing a sequence of query images sampled from a video. 
Hence, temporally close frames can be expected to show spatial coherence in their pose estimates. 
However, due to the difference in appearance between objects that are present in the scenes and the objects that are used for the attack, as well as due to viewpoint changes, there can be many outlier poses. 
Still, Algorithm 1 (of the main paper) is 
able to identify subsets of camera poses that allow to appropriately position the 3D models of the objects, demonstrating the robustness of the approach. 

\PAR{Failure cases.} 
For each scene, failure cases of the alignment step are highlighted using red colored boxes. 
Alignments that result in positioning the attacking object such that it is either too far from the corresponding object in the scene or its "up-direction" is very different from that of the corresponding object are considered as failure cases. 
Only visual inspection has been used to decide whether a case is considered 
a success or failure.
As can be seen from visualizations, the 3D models of the attacking objects that result in failures are often quite different (in terms of appearance) from the corresponding objects in the scene. 
Some failure cases such as the \textit{Sofa Linanas} in Figure \ref{fig:ikea_scene05_align2} can be attributed to this 
difference while many other cases such as \textit{Sofa Soderhamn} in Figure \ref{fig:ikea_scene04_align2} and \textit{Stool Kyrre} in Figure \ref{fig:ikea_scene05_align2} show the success in such difficult cases.
Other reasons for failure are a low number of matches from texture-less or weakly textured objects. \textit{Chair Odger} in Figure \ref{fig:ikea_scene04_align2} and \textit{Chair Froset} in Figure \ref{fig:ikea_scene01_align2} (using Superpoint features~\cite{detone2017superpoint} with the Superglue matcher~\cite{sarlin20superglue}), 
are examples of such cases.
Scene02, Scene04, and Scene06, shown in Figures \ref{fig:ikea_scene02_overview}-\ref{fig:ikea_scene02_align2}, \ref{fig:ikea_scene04_overview}-\ref{fig:ikea_scene04_align2}, \ref{fig:ikea_scene06_overview}-\ref{fig:ikea_scene06_align2} respectively, are complex rooms (\eg, an open concept kitchen in \textit{Scene04}) composed of very similar looking objects, and hence challenging cases for such an attack. This results in more failure cases. 
Still, as shown in the figures, the attack succeeds in many cases, which shows its feasibility. 


\section{RIO10 example alignment results}\label{sec:qual_rio}
\begin{figure*}[t!]
    \centering
    \includegraphics[width=0.55\linewidth]{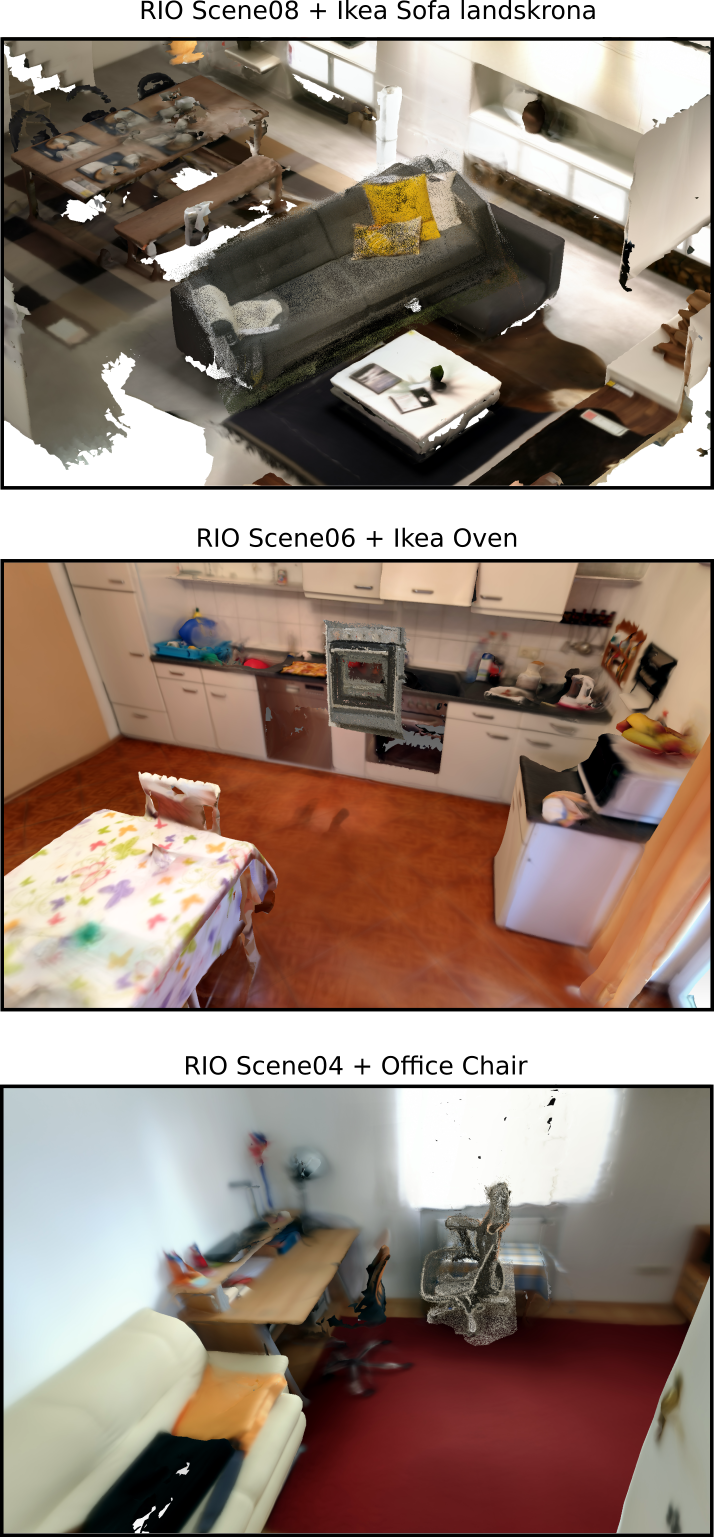}
    \caption{Alignment results for querying scenes from the RIO10 dataset~\cite{wald2020} with objects from the Office-Objects and IKEA-Objects datasets (\cf Fig. 6 in the main paper for quantitative results). 
    The textured mesh corresponds to the scene and the point cloud corresponds to the 3D model of the object used for the attack. The alignment was produced using the poses obtained from the server, using Superpoint+Superglue for matching. As can be seen, it is possible to position the objects with reasonable accuracy, despite differences in appearance and geometry.} 
    \label{fig:rio_qual}
\end{figure*}

\begin{figure*}[t!]
    \centering
    \includegraphics[width=0.98\linewidth]{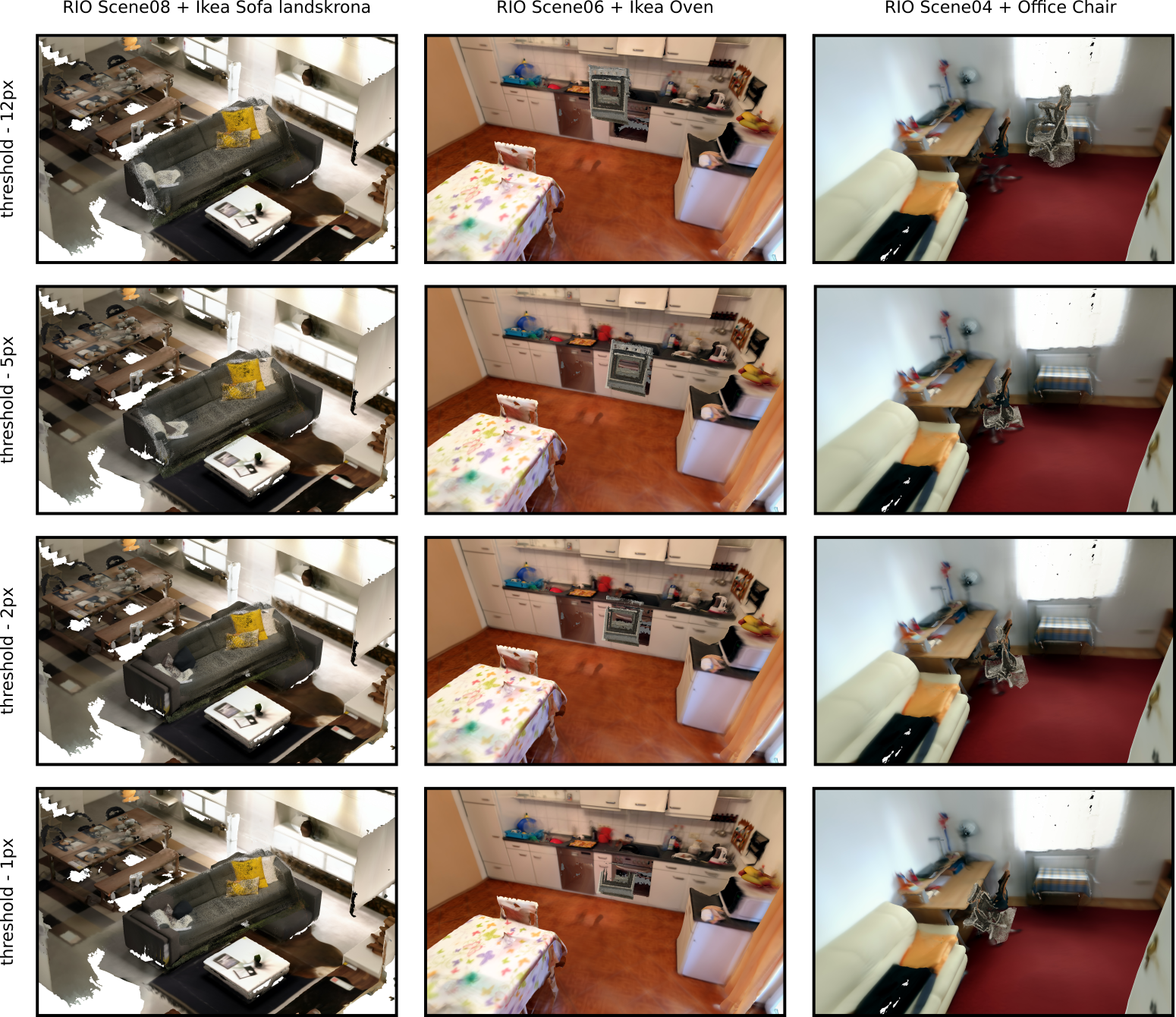}
    \caption{Impact of changing the inlier threshold for RANSAC based localization on alignment - qualitative results. The final aligned object is not impacted drastically in most cases by a stricter RANSAC threshold. In fact, in several cases, it results in a better alignment of the attacking object with the corresponding object in the scene. 
    The textured mesh corresponds to the scene and the point cloud corresponds to the 3D model of the object used for the attack. The alignment was produced using the poses obtained from the server, using Superpoint+Superglue for matching} 
    \label{fig:thresh_exp}
\end{figure*}

In Section 6 of the main paper, we discuss a potential strategy to defend against the attack introduced in our work. 
Yet, Figure 6 of the main paper shows that this strategy causes the localization process to not only reject malicious queries, but also to reject genuine query images. 
In Figure 6 of the main paper, we consider 3 different objects attacking 3 different scenes of the RIO10 dataset. 
To visualize these scenarios, 
Figure \ref{fig:rio_qual} shows qualitative results for object alignments in scenes corresponding to those plots.
These results further emphasize that the attack does not require images of the exact same objects as present in the scene, but can also be carried out using images of similar instances from the same class of objects. Furthermore, these alignments also do not change significantly when the HLoc\cite{sarlin2019coarse} based server uses stricter inlier thresholds (5,2 or 1 pixel as compared to the default 12 pixels) for the RANSAC based localization process. This counters another simple defence strategy of using stricter inlier thresholds in such a localization pipeline to avoid localization of malicious queries. Figure \ref{fig:thresh_exp} shows the qualitative results of object alignment for varying inlier thresholds. As can be seen, the alignment does not get worse (in fact, improves) upon using stricter thresholds for the three cases considered here. Note that for these examples, although the number of inliers decreases as the threshold become stricter, the solutions of poses with least sum of reprojection errors are still able to position the attack-object appropriately. A more rigorous experimentation against this defence would definitely be more insightful but that would require an appropriate quantitative measure of the quality of alignment and this is left as future work.  

\section{Additional results for object presence classification}
For the task object presence classification, we present precision and recall results for each of the objects in \textit{IKEA-Objects} in Table \ref{tab:prec_recall_obj}.


\begin{table*}[t!]
\centering
\begin{tabular}{|l|llllll|llllll|}
\hline
\multicolumn{1}{|c|}{\multirow{3}{*}{\textbf{Object Name}}} & \multicolumn{6}{c|}{\textbf{Superpoint + Superglue}}                                                                                                & \multicolumn{6}{c|}{\textbf{R2D2 + NN}}                                                                                                             \\ \cline{2-13} 
\multicolumn{1}{|c|}{}                                      & \multicolumn{2}{l|}{$10^{\circ}$,0.25m}              & \multicolumn{2}{l|}{$30^{\circ}$, 0.5m}              & \multicolumn{2}{l|}{$60^{\circ}$, 2m} & \multicolumn{2}{l|}{$10^{\circ}$,0.25m}              & \multicolumn{2}{l|}{$30^{\circ}$, 0.5m}              & \multicolumn{2}{l|}{$60^{\circ}$, 2m} \\ \cline{2-13} 
\multicolumn{1}{|c|}{}                                      & \multicolumn{1}{l|}{P}    & \multicolumn{1}{l|}{R}   & \multicolumn{1}{l|}{P}    & \multicolumn{1}{l|}{R}   & \multicolumn{1}{l|}{P}       & R      & \multicolumn{1}{l|}{P}    & \multicolumn{1}{l|}{R}   & \multicolumn{1}{l|}{P}    & \multicolumn{1}{l|}{R}   & \multicolumn{1}{l|}{P}       & R      \\ \hline
bookshelf                                                   & \multicolumn{1}{l|}{1}    & \multicolumn{1}{l|}{0.5} & \multicolumn{1}{l|}{0.5}  & \multicolumn{1}{l|}{0.5} & \multicolumn{1}{l|}{1}       & 0.5    & \multicolumn{1}{l|}{0.5}  & \multicolumn{1}{l|}{0.5} & \multicolumn{1}{l|}{1}    & \multicolumn{1}{l|}{0.5} & \multicolumn{1}{l|}{0.5}     & 0.5    \\ \hline
chair\_agam                                                 & \multicolumn{1}{l|}{0}    & \multicolumn{1}{l|}{0}   & \multicolumn{1}{l|}{0}    & \multicolumn{1}{l|}{0}   & \multicolumn{1}{l|}{0}       & 0      & \multicolumn{1}{l|}{0}    & \multicolumn{1}{l|}{0}   & \multicolumn{1}{l|}{0}    & \multicolumn{1}{l|}{0}   & \multicolumn{1}{l|}{0}       & 0      \\ \hline
chair\_froset                                               & \multicolumn{1}{l|}{0}    & \multicolumn{1}{l|}{0}   & \multicolumn{1}{l|}{0}    & \multicolumn{1}{l|}{0}   & \multicolumn{1}{l|}{0}       & 0      & \multicolumn{1}{l|}{0}    & \multicolumn{1}{l|}{0}   & \multicolumn{1}{l|}{0}    & \multicolumn{1}{l|}{0}   & \multicolumn{1}{l|}{0}       & 0      \\ \hline
chair\_gaming                                               & \multicolumn{1}{l|}{0.33} & \multicolumn{1}{l|}{1}   & \multicolumn{1}{l|}{0.5}  & \multicolumn{1}{l|}{1}   & \multicolumn{1}{l|}{0}       & 0      & \multicolumn{1}{l|}{1}    & \multicolumn{1}{l|}{1}   & \multicolumn{1}{l|}{1}    & \multicolumn{1}{l|}{1}   & \multicolumn{1}{l|}{0.5}     & 1      \\ \hline
chair\_linneback                                            & \multicolumn{1}{l|}{0.5}  & \multicolumn{1}{l|}{1}   & \multicolumn{1}{l|}{0.5}  & \multicolumn{1}{l|}{1}   & \multicolumn{1}{l|}{0.5}     & 1      & \multicolumn{1}{l|}{0.5}  & \multicolumn{1}{l|}{1}   & \multicolumn{1}{l|}{0.5}  & \multicolumn{1}{l|}{1}   & \multicolumn{1}{l|}{0.5}     & 1      \\ \hline
chair\_odger                                                & \multicolumn{1}{l|}{0.33} & \multicolumn{1}{l|}{0.5} & \multicolumn{1}{l|}{0.33} & \multicolumn{1}{l|}{0.5} & \multicolumn{1}{l|}{0.5}     & 0.5    & \multicolumn{1}{l|}{1}    & \multicolumn{1}{l|}{0.5} & \multicolumn{1}{l|}{1}    & \multicolumn{1}{l|}{0.5} & \multicolumn{1}{l|}{0.5}     & 0.5    \\ \hline
chair\_poang\_small                                         & \multicolumn{1}{l|}{1}    & \multicolumn{1}{l|}{1}   & \multicolumn{1}{l|}{1}    & \multicolumn{1}{l|}{1}   & \multicolumn{1}{l|}{1}       & 1      & \multicolumn{1}{l|}{1}    & \multicolumn{1}{l|}{1}   & \multicolumn{1}{l|}{1}    & \multicolumn{1}{l|}{1}   & \multicolumn{1}{l|}{1}       & 1      \\ \hline
chair\_storsele                                             & \multicolumn{1}{l|}{0.33} & \multicolumn{1}{l|}{1}   & \multicolumn{1}{l|}{0.33} & \multicolumn{1}{l|}{1}   & \multicolumn{1}{l|}{1}       & 1      & \multicolumn{1}{l|}{0.5}  & \multicolumn{1}{l|}{1}   & \multicolumn{1}{l|}{0.25} & \multicolumn{1}{l|}{1}   & \multicolumn{1}{l|}{0.5}     & 1      \\ \hline
chair\_strandmon                                            & \multicolumn{1}{l|}{0}    & \multicolumn{1}{l|}{0}   & \multicolumn{1}{l|}{0}    & \multicolumn{1}{l|}{0}   & \multicolumn{1}{l|}{0}       & 0      & \multicolumn{1}{l|}{0.5}  & \multicolumn{1}{l|}{1}   & \multicolumn{1}{l|}{0.5}  & \multicolumn{1}{l|}{1}   & \multicolumn{1}{l|}{0.5}     & 1      \\ \hline
chair\_vedbo                                                & \multicolumn{1}{l|}{1}    & \multicolumn{1}{l|}{1}   & \multicolumn{1}{l|}{1}    & \multicolumn{1}{l|}{1}   & \multicolumn{1}{l|}{0.5}     & 1      & \multicolumn{1}{l|}{0.5}  & \multicolumn{1}{l|}{1}   & \multicolumn{1}{l|}{0.5}  & \multicolumn{1}{l|}{1}   & \multicolumn{1}{l|}{0.5}     & 1      \\ \hline
cupboard\_hauga                                             & \multicolumn{1}{l|}{1}    & \multicolumn{1}{l|}{1}   & \multicolumn{1}{l|}{1}    & \multicolumn{1}{l|}{1}   & \multicolumn{1}{l|}{1}       & 1      & \multicolumn{1}{l|}{1}    & \multicolumn{1}{l|}{1}   & \multicolumn{1}{l|}{1}    & \multicolumn{1}{l|}{1}   & \multicolumn{1}{l|}{1}       & 1      \\ \hline
cupboard\_kallax                                            & \multicolumn{1}{l|}{1}    & \multicolumn{1}{l|}{1}   & \multicolumn{1}{l|}{1}    & \multicolumn{1}{l|}{1}   & \multicolumn{1}{l|}{1}       & 1      & \multicolumn{1}{l|}{0}    & \multicolumn{1}{l|}{0}   & \multicolumn{1}{l|}{0}    & \multicolumn{1}{l|}{0}   & \multicolumn{1}{l|}{0.25}    & 1      \\ \hline
cycle                                                       & \multicolumn{1}{l|}{0}    & \multicolumn{1}{l|}{0}   & \multicolumn{1}{l|}{0}    & \multicolumn{1}{l|}{0}   & \multicolumn{1}{l|}{0}       & 0      & \multicolumn{1}{l|}{0.5}  & \multicolumn{1}{l|}{1}   & \multicolumn{1}{l|}{0.5}  & \multicolumn{1}{l|}{1}   & \multicolumn{1}{l|}{0.5}     & 1      \\ \hline
klipsk\_bed\_table                                          & \multicolumn{1}{l|}{1}    & \multicolumn{1}{l|}{1}   & \multicolumn{1}{l|}{1}    & \multicolumn{1}{l|}{1}   & \multicolumn{1}{l|}{1}       & 1      & \multicolumn{1}{l|}{1}    & \multicolumn{1}{l|}{1}   & \multicolumn{1}{l|}{1}    & \multicolumn{1}{l|}{1}   & \multicolumn{1}{l|}{0.5}     & 1      \\ \hline
lamp\_ceiling\_agunarryd                                    & \multicolumn{1}{l|}{1}    & \multicolumn{1}{l|}{1}   & \multicolumn{1}{l|}{1}    & \multicolumn{1}{l|}{1}   & \multicolumn{1}{l|}{1}       & 1      & \multicolumn{1}{l|}{0.33} & \multicolumn{1}{l|}{1}   & \multicolumn{1}{l|}{0}    & \multicolumn{1}{l|}{0}   & \multicolumn{1}{l|}{0.33}    & 1      \\ \hline
lamp\_ceiling\_appleviken                                   & \multicolumn{1}{l|}{1}    & \multicolumn{1}{l|}{1}   & \multicolumn{1}{l|}{1}    & \multicolumn{1}{l|}{1}   & \multicolumn{1}{l|}{1}       & 1      & \multicolumn{1}{l|}{0}    & \multicolumn{1}{l|}{0}   & \multicolumn{1}{l|}{0}    & \multicolumn{1}{l|}{0}   & \multicolumn{1}{l|}{0}       & 0      \\ \hline
lamp\_ceiling\_mojna                                        & \multicolumn{1}{l|}{0}    & \multicolumn{1}{l|}{0}   & \multicolumn{1}{l|}{0}    & \multicolumn{1}{l|}{0}   & \multicolumn{1}{l|}{0}       & 0      & \multicolumn{1}{l|}{0}    & \multicolumn{1}{l|}{0}   & \multicolumn{1}{l|}{0}    & \multicolumn{1}{l|}{0}   & \multicolumn{1}{l|}{0}       & 0      \\ \hline
lamp\_ceiling\_nymane                                       & \multicolumn{1}{l|}{0}    & \multicolumn{1}{l|}{0}   & \multicolumn{1}{l|}{1}    & \multicolumn{1}{l|}{0.5} & \multicolumn{1}{l|}{0.5}     & 0.5    & \multicolumn{1}{l|}{1}    & \multicolumn{1}{l|}{0.5} & \multicolumn{1}{l|}{0}    & \multicolumn{1}{l|}{0}   & \multicolumn{1}{l|}{0.25}    & 0.5    \\ \hline
lamp\_ceiling\_ranarp                                       & \multicolumn{1}{l|}{0}    & \multicolumn{1}{l|}{0}   & \multicolumn{1}{l|}{0}    & \multicolumn{1}{l|}{0}   & \multicolumn{1}{l|}{0}       & 0      & \multicolumn{1}{l|}{0}    & \multicolumn{1}{l|}{0}   & \multicolumn{1}{l|}{0}    & \multicolumn{1}{l|}{0}   & \multicolumn{1}{l|}{0}       & 0      \\ \hline
lamp\_evedal                                                & \multicolumn{1}{l|}{1}    & \multicolumn{1}{l|}{1}   & \multicolumn{1}{l|}{1}    & \multicolumn{1}{l|}{1}   & \multicolumn{1}{l|}{1}       & 1      & \multicolumn{1}{l|}{1}    & \multicolumn{1}{l|}{1}   & \multicolumn{1}{l|}{1}    & \multicolumn{1}{l|}{1}   & \multicolumn{1}{l|}{1}       & 1      \\ \hline
lamp\_fancy                                                 & \multicolumn{1}{l|}{1}    & \multicolumn{1}{l|}{1}   & \multicolumn{1}{l|}{1}    & \multicolumn{1}{l|}{1}   & \multicolumn{1}{l|}{1}       & 1      & \multicolumn{1}{l|}{1}    & \multicolumn{1}{l|}{1}   & \multicolumn{1}{l|}{1}    & \multicolumn{1}{l|}{1}   & \multicolumn{1}{l|}{1}       & 1      \\ \hline
lamp\_navlinge                                              & \multicolumn{1}{l|}{0.33} & \multicolumn{1}{l|}{0.5} & \multicolumn{1}{l|}{0.5}  & \multicolumn{1}{l|}{0.5} & \multicolumn{1}{l|}{0.5}     & 0.5    & \multicolumn{1}{l|}{0.25} & \multicolumn{1}{l|}{0.5} & \multicolumn{1}{l|}{0.33} & \multicolumn{1}{l|}{0.5} & \multicolumn{1}{l|}{0.25}    & 0.5    \\ \hline
lamp\_star                                                  & \multicolumn{1}{l|}{0}    & \multicolumn{1}{l|}{0}   & \multicolumn{1}{l|}{0}    & \multicolumn{1}{l|}{0}   & \multicolumn{1}{l|}{0}       & 0      & \multicolumn{1}{l|}{1}    & \multicolumn{1}{l|}{1}   & \multicolumn{1}{l|}{1}    & \multicolumn{1}{l|}{1}   & \multicolumn{1}{l|}{1}       & 1      \\ \hline
lamp\_table\_misterhult                                     & \multicolumn{1}{l|}{1}    & \multicolumn{1}{l|}{1}   & \multicolumn{1}{l|}{1}    & \multicolumn{1}{l|}{1}   & \multicolumn{1}{l|}{1}       & 1      & \multicolumn{1}{l|}{0}    & \multicolumn{1}{l|}{0}   & \multicolumn{1}{l|}{0}    & \multicolumn{1}{l|}{0}   & \multicolumn{1}{l|}{0}       & 0      \\ \hline
lamp\_table\_nymane                                         & \multicolumn{1}{l|}{0.5}  & \multicolumn{1}{l|}{0.5} & \multicolumn{1}{l|}{1}    & \multicolumn{1}{l|}{0.5} & \multicolumn{1}{l|}{0.33}    & 0.5    & \multicolumn{1}{l|}{1}    & \multicolumn{1}{l|}{0.5} & \multicolumn{1}{l|}{1}    & \multicolumn{1}{l|}{0.5} & \multicolumn{1}{l|}{1}       & 0.5    \\ \hline
lamp\_table\_tertial                                        & \multicolumn{1}{l|}{0}    & \multicolumn{1}{l|}{0}   & \multicolumn{1}{l|}{0.5}  & \multicolumn{1}{l|}{1}   & \multicolumn{1}{l|}{1}       & 1      & \multicolumn{1}{l|}{0.5}  & \multicolumn{1}{l|}{1}   & \multicolumn{1}{l|}{0}    & \multicolumn{1}{l|}{0}   & \multicolumn{1}{l|}{0.33}    & 1      \\ \hline
organizer\_kvarnik                                          & \multicolumn{1}{l|}{0.5}  & \multicolumn{1}{l|}{1}   & \multicolumn{1}{l|}{0}    & \multicolumn{1}{l|}{0}   & \multicolumn{1}{l|}{0}       & 0      & \multicolumn{1}{l|}{1}    & \multicolumn{1}{l|}{1}   & \multicolumn{1}{l|}{1}    & \multicolumn{1}{l|}{1}   & \multicolumn{1}{l|}{0.5}     & 1      \\ \hline
oven                                                        & \multicolumn{1}{l|}{1}    & \multicolumn{1}{l|}{0.5} & \multicolumn{1}{l|}{1}    & \multicolumn{1}{l|}{0.5} & \multicolumn{1}{l|}{1}       & 0.5    & \multicolumn{1}{l|}{1}    & \multicolumn{1}{l|}{0.5} & \multicolumn{1}{l|}{1}    & \multicolumn{1}{l|}{0.5} & \multicolumn{1}{l|}{0.5}     & 0.5    \\ \hline
sofa\_landskrona                                            & \multicolumn{1}{l|}{1}    & \multicolumn{1}{l|}{1}   & \multicolumn{1}{l|}{1}    & \multicolumn{1}{l|}{1}   & \multicolumn{1}{l|}{0.5}     & 1      & \multicolumn{1}{l|}{1}    & \multicolumn{1}{l|}{1}   & \multicolumn{1}{l|}{1}    & \multicolumn{1}{l|}{1}   & \multicolumn{1}{l|}{1}       & 1      \\ \hline
sofa\_linanas                                               & \multicolumn{1}{l|}{0}    & \multicolumn{1}{l|}{0}   & \multicolumn{1}{l|}{0}    & \multicolumn{1}{l|}{0}   & \multicolumn{1}{l|}{0}       & 0      & \multicolumn{1}{l|}{0}    & \multicolumn{1}{l|}{0}   & \multicolumn{1}{l|}{0}    & \multicolumn{1}{l|}{0}   & \multicolumn{1}{l|}{0}       & 0      \\ \hline
sofa\_soderhamn                                             & \multicolumn{1}{l|}{0}    & \multicolumn{1}{l|}{0}   & \multicolumn{1}{l|}{0}    & \multicolumn{1}{l|}{0}   & \multicolumn{1}{l|}{0}       & 0      & \multicolumn{1}{l|}{0}    & \multicolumn{1}{l|}{0}   & \multicolumn{1}{l|}{0}    & \multicolumn{1}{l|}{0}   & \multicolumn{1}{l|}{0}       & 0      \\ \hline
stool\_kyrre                                                & \multicolumn{1}{l|}{1}    & \multicolumn{1}{l|}{1}   & \multicolumn{1}{l|}{1}    & \multicolumn{1}{l|}{1}   & \multicolumn{1}{l|}{1}       & 1      & \multicolumn{1}{l|}{1}    & \multicolumn{1}{l|}{1}   & \multicolumn{1}{l|}{1}    & \multicolumn{1}{l|}{1}   & \multicolumn{1}{l|}{1}       & 1      \\ \hline
stool\_marius                                               & \multicolumn{1}{l|}{1}    & \multicolumn{1}{l|}{1}   & \multicolumn{1}{l|}{1}    & \multicolumn{1}{l|}{1}   & \multicolumn{1}{l|}{1}       & 1      & \multicolumn{1}{l|}{1}    & \multicolumn{1}{l|}{1}   & \multicolumn{1}{l|}{1}    & \multicolumn{1}{l|}{1}   & \multicolumn{1}{l|}{1}       & 1      \\ \hline
strainer                                                    & \multicolumn{1}{l|}{0}    & \multicolumn{1}{l|}{0}   & \multicolumn{1}{l|}{0}    & \multicolumn{1}{l|}{0}   & \multicolumn{1}{l|}{0}       & 0      & \multicolumn{1}{l|}{0}    & \multicolumn{1}{l|}{0}   & \multicolumn{1}{l|}{0}    & \multicolumn{1}{l|}{0}   & \multicolumn{1}{l|}{0}       & 0      \\ \hline
table\_corner\_gladom                                       & \multicolumn{1}{l|}{1}    & \multicolumn{1}{l|}{1}   & \multicolumn{1}{l|}{1}    & \multicolumn{1}{l|}{1}   & \multicolumn{1}{l|}{1}       & 1      & \multicolumn{1}{l|}{0.5}  & \multicolumn{1}{l|}{1}   & \multicolumn{1}{l|}{0.5}  & \multicolumn{1}{l|}{1}   & \multicolumn{1}{l|}{0.5}     & 1      \\ \hline
table\_lisabo\_square                                       & \multicolumn{1}{l|}{1}    & \multicolumn{1}{l|}{0.5} & \multicolumn{1}{l|}{0.5}  & \multicolumn{1}{l|}{0.5} & \multicolumn{1}{l|}{0.25}    & 0.5    & \multicolumn{1}{l|}{1}    & \multicolumn{1}{l|}{0.5} & \multicolumn{1}{l|}{1}    & \multicolumn{1}{l|}{0.5} & \multicolumn{1}{l|}{1}       & 0.5    \\ \hline
vas\_gradvis                                                & \multicolumn{1}{l|}{0.25} & \multicolumn{1}{l|}{1}   & \multicolumn{1}{l|}{1}    & \multicolumn{1}{l|}{1}   & \multicolumn{1}{l|}{0.5}     & 1      & \multicolumn{1}{l|}{0}    & \multicolumn{1}{l|}{0}   & \multicolumn{1}{l|}{0}    & \multicolumn{1}{l|}{0}   & \multicolumn{1}{l|}{0.5}     & 1      \\ \hline
wall\_hanging\_crescent                                     & \multicolumn{1}{l|}{0.33} & \multicolumn{1}{l|}{0.5} & \multicolumn{1}{l|}{0.25} & \multicolumn{1}{l|}{0.5} & \multicolumn{1}{l|}{0.33}    & 0.5    & \multicolumn{1}{l|}{0}    & \multicolumn{1}{l|}{0}   & \multicolumn{1}{l|}{1}    & \multicolumn{1}{l|}{0.5} & \multicolumn{1}{l|}{0.5}     & 0.5    \\ \hline
\end{tabular}
\caption{Object wise precision and recall results for \textit{IKEA-Objects} when attacking \textit{IKEA-Scenes}}
\label{tab:prec_recall_obj}
\end{table*}

\end{document}